\documentclass[authoryear,1p]{elsarticle}
\usepackage[colorlinks=true,linkcolor=black, citecolor=blue, urlcolor=blue]{hyperref}

\usepackage{amssymb}
\usepackage{adjustbox}
\usepackage{multirow}

\usepackage[inline]{enumitem}
\newlist{inline}{enumerate*}{1}
\setlist[inline]{label=(\roman*)}

\usepackage{amsmath}

\journal{Transport Research Part C}

\begin{document}

\begin{frontmatter}



\title{Synthesising Activity Participations and Scheduling with Deep Generative Machine Learning}


\author[inst1]{Fred Shone}

\author[inst1]{Tim Hillel}

\affiliation[inst1]{organization={University College London (UCL), Behaviour and Infrastructure Group (BIG)},%Department and Organization
            addressline={Chadwick Building, Gower Street }, 
            city={London},
            postcode={WC1E 6BT}, 
            country={United Kingdom}}

\begin{abstract}
Using a deep generative machine learning approach, we synthesise human activity participations and scheduling; i.e. the choices of what activities to participate in and when. Activity schedules are a core component of many applied transport, energy, and epidemiology models. Our data-driven approach directly learns the distributions resulting from human preferences and scheduling logic without the need for complex interacting combinations of sub-models and custom rules. This makes our approach significantly faster and simpler to operate than existing approaches to synthesise or anonymise schedule data. We additionally contribute a novel schedule representation and a comprehensive evaluation framework. We evaluate a range of schedule encoding and deep model architecture combinations. The evaluation shows our approach can rapidly generate large, diverse, novel, and realistic synthetic samples of activity schedules.

\end{abstract}

% \centering
% \includegraphics[width=0.5\linewidth]{images/summary.png}

\begin{highlights}
    \item We contribute a novel approach to synthesising human activity schedules using deep generative modelling. Specifically, we demonstrate the rapid synthesis of large, realistic, and representative samples of human activity schedules using Variational Auto-Encoders.
    \item We propose an activity schedule encoding, using a continuous encoding of activity durations.
    \item  We present a comprehensive framework for evaluating the quality of model outputs, focusing on distribution-level factors which are crucial for simulations. We use human interpretable distance metrics such that users can reasonably assess the suitability of models for various tasks.
\end{highlights}


\end{frontmatter}


\section{Introduction}
\label{sec:intro}
We propose a novel approach to an existing modelling challenge - the generation or \emph{synthesis} of realistic samples of activity schedules. We define activity schedules as 24-hour-long sequences of activity type participations belonging to an individual, with associated start times and durations for each activity. Activity schedules are critical for many domains concerned with predicting or simulating human behaviour, such as activity-based models and agent-based simulations, including MATSim \citep{MATSim}. In both cases, large samples of activity schedules are used to represent patterns of activities for representative populations of agents. Such models are used as decision-support tools for transport, energy and epidemiological scenarios.

Our definition of activity schedules can be decomposed into:
\begin{inline}
    \item  participations - \textit{if or how many times an individual chooses to undertake certain activities}, and 
    \item timings - \textit{when and for how long to undertake the chosen activities}.
\end{inline}
Note, we do not consider here other aspects such as activity locations or associated trips and their attributes, such as mode choices.

The participation and timing of multiple activities for an individual creates a high-dimensional object with complex joint distributions. The complex distributions of activity schedules are the result of the real-world processes by which people plan and undertake their days. This process includes the consideration of
\begin{inline}
    \item physical and temporal constraints, such as not being able to partake in multiple activities at the same time or fitting all activities into a finite time budget, 
    \item personal and social constraints, such as arriving at work on time or needing to buy food, 
    \item individual preferences, such as the desire to participate in leisure activities or preferences towards different activity start-times, and 
    \item interactions with others, such as coordinating shared activities in a household or via congestion.
\end{inline}

The prevailing approach in both research and practice is to decompose the scheduling process into a series of discrete choices, applied sequentially. Each of these choices is conditioned on observed information, such as socio-economic attributes, and on previously made choices. To ensure temporal consistency, choices are then combined with rule-based scheduling algorithms. We highlight three main critiques of this approach:
\begin{inline}
    \item sequential choices presume some order of decision making or causation that may be unrealistic, 
    \item the combination of discrete choices and rules is simplified such that it cannot reproduce the real diversity of observed activity schedules, and 
    \item the complex combination of multiple interacting sub-models and rules is challenging to develop, calibrate and use.
\end{inline}

We demonstrate a novel approach for activity schedule synthesis using deep generative machine learning (ML). Our approach allows for the simultaneous synthesis of schedules, combining choices that would typically be made separately. Additionally, deep generative ML explicitly models the variation in observed data, without the need for conditional information. This results in three key benefits:
\begin{inline} 
    \item simultaneous synthesis potentially allows more realistic interaction of choice components,
    \item deep generative ML better models the distribution of output sequences, and 
    \item in combination, the approach is simple and fast to apply. 
\end{inline}
We comprehensively evaluate the quality of output activity schedules both individually and in aggregate.

We show that by incorporating a generative approach, we can better model the true distribution, or variance, of real schedules compared to existing conditional supervised learning approaches. However, a major limitation is that we cannot model conditionality, such as in a typical behavioural modelling framework. For example, we cannot model responses to changes in socio-economic attributes or accessibility. This limits the application to where conditionality is not required, such as:
\begin{inline} 
    \item synthesis, 
    \item data anonymisation, and 
    \item diverse up-sampling.
\end{inline}

Ultimately, this work is intended to demonstrate and establish generative modelling approaches in this domain, leaving their introduction into models and frameworks with conditionality to future work. We share experimentation with data encodings and model architectures to assist future approaches using deep ML and generative approaches. Code for this research is available from the Caveat GitHub repository\footnote{https://github.com/big-ucl/caveat}.

\subsection{Activity Scheduling}

Table~\ref{tab:models} provides an overview of existing applied modelling frameworks that incorporate human activity scheduling. These approaches all make use of combinations of bespoke rules and statistical choice models to model schedules. These approaches seek to realistically model the activity scheduling process through the incorporation of realistic demand representation and generation processes. The incorporation of more realism has led to complex systems of interacting model components. Generally, these models are slow and expensive to develop. For example, they typically require the collection of data describing all possible realisations of choice sets and significant human effort to calibrate.

\begin{table}
    \footnotesize
    \caption{Summary of Existing Activity Scheduling Approaches}
    \vspace{2ex}
    \centering
        \begin{tabular}{l c c }
        \hline
        Model/Framework & Activity Participation & Activity Timing \\
        \hline
        \hline
        TASHA \citep{TASHA} & Rule-based & Rule-based \\
        ALBATROSS \citep{ALBATROSS} & Rule-based & Rule-based \\
        FAMOS \citep{FAMOS} & Nested-logit models & Hazard models \\
        CEMPDAP \citep{CEMDAP} & Nested-logit models & Hazard models \\
        ADAPTS \citep{ADAPTS} & Rule-based & Rule-based \\
        DaySim \citep{DaySim} & Multinomial-logit models & Multinomial-logit models \\
        SDS \citep{SDS} & Markov Chain Monte Carlo & Rule-based \\
        \hline
    \end{tabular}           
    \label{tab:models}
\end{table}

A more theoretical critique of the above approaches is that they decompose schedule modelling into sequential choices. This presumes some order of decision-making that may be unrealistic. It is not clear, for example, if a person chooses to go shopping after work because they finished early, or if they finished early to go shopping. A common response to this is to combine discrete choices within joint models, such that they are modelled simultaneously.

\cite{POUGALA2023104291} combine activity participation and scheduling into a simultaneous model. They achieve this via a simulated and sampled choice set. This approach is consistent with existing behavioural theory and is flexible enough to be extended to a variety of problems, constraints, and utility specifications. However, both estimation of the parameters and simulation of schedules are computationally expensive, limiting scalability. \cite{Manser2021ResolvingTS} manage to scale the approach to application as part of an activity-based transport demand model, but they limit the scope of the simultaneous approach to activity timings only.

A core critique of existing approaches is that they under-represent the diversity of real behaviours. For example, tour-based approaches, such as used by \cite{miller2005tour}, restrict schedule activity participations to a reduced set of common patterns. By omitting less common behavioural patterns, these models introduce biases that limit their realism. Machine Learning (ML) approaches, such as used by \cite{koushikActivityScheduleModeling2023} demonstrate the simultaneous modelling of participations and scheduling. However, by taking a discriminative approach, the model effectively generates the most likely activity sequence for each demographic group, therefore ignoring realistic variation within the groups. Such simplifications limit the realism of downstream models or simulations.

Activity schedules have also been treated purely as sequences with probabilistic transitions. For example, \cite{pHMM} model activity sequences as Hidden Markov Models. Their approach allows consideration of both frequent and infrequent activities and resulting sequences. However, their work does not consider activity durations and is limited to the analysis and clustering of data rather than the generation of new sequences of activities.

\subsection{Supervised Discriminative Machine Learning Approaches to Activity Scheduling}
\label{sec:mlscheduling}

Traditional approaches to discrete choice modelling are structured such that their parameters can be both constrained and interpreted to be consistent with behavioural theories. However, increasingly researchers are applying purely data-driven ML approaches to these problems, including for schedule prediction.

\cite{dap_ml} consider activity sequences (participations) across multiple days as a classification problem; comparing the performance of various bagging and boosting models for this task. They simplify the activity scheduling problem by identifying a reduced set of common day tours and then treating scheduling as a supervised discriminative (or conditional) learning problem. \cite{trees} use Random Forests to predict activity start times and durations given person attributes. They then use heuristic rules to assemble these activity times and durations into schedules. Time is discretised into various bin sizes from 10 to 360 minutes. In such discriminative ML approaches, evaluation is made by reporting classification accuracy for various predefined clusters against ground-truth data. The use of accuracy for evaluation makes it difficult to assess the impact of the approaches on a broader scheduling framework.

The above ML approaches tackle different components of a typical activity modelling framework - activity participations and timing.  However, common to many traditional approaches, choices are discretised into simplified discrete sets, limiting realism, and requiring schedules to be assembled via multiple interaction models or heuristics.

\cite{koushikActivityScheduleModeling2023} use a deep ML model to generate activity schedules for given agent attributes using recurrent neural networks (RNNs). Their model considers activity participations and their scheduling jointly as a simultaneous choice. They discretise 24-hour schedules into five-minute steps and consider nine different types of activity. They sample activities at each step based on maximum likelihood estimation, but do not include sampling in the inference process. As such, we do not consider this to be a generative model. Evaluation is made through the consideration of key selected marginal distributions, including the distribution of start times by type of activity. The authors find replicating aggregate realism challenging, particularly correctly representing short and infrequently observed activities. There is limited consideration of individual sample quality, or more broadly, of the distribution of their results.

In all cases, these models take a discriminative or conditional approach, where variation in observed schedules must be described by either explanatory variables or variation in arbitrary error terms (typically normal). Our work provides an alternative to the above by approximating the joint distribution of schedule activity participations and timings, and by explicitly synthesising variation by drawing from this joint distribution using a generative approach. This removes the need for explanatory data to explain observed variation.

\subsection{Activity Schedule Evaluation}

Generated activity schedule evaluation is typically performed through the comparison of aggregate distributions with a target sample of schedules. The models presented in Table~\ref{tab:models} are primarily concerned with the representation of travel demand and therefore focus on the extraction and comparison of distributions such as total trip rates.

More recently, \cite{POUGALA2023104291} and \cite{drchalVALFRAMValidationFramework2016} make a more holistic evaluation through consideration of multiple marginal and conditioned distributions, such as different activity participation rates and start times.

\begin{figure}
    \centering
    \includegraphics[trim={0 0.2cm 0 1cm}, clip, width=1\linewidth]{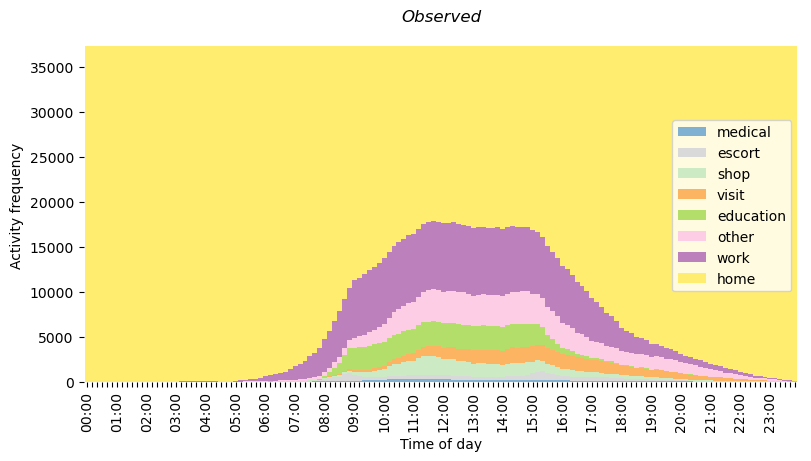}
    \caption{UK National Travel Survey (2023) Activity Frequencies: Aggregate activity participation by 10-minute time bin}
    \label{fig:nts_freq}
\end{figure}

A common evaluation approach involves analysing aggregate activity participation rates by time bin. Figure~\ref{fig:nts_freq} shows this distribution for 2023 UK National Travel Survey data. Such evaluation considers total or relative participation in activities by time of day. However, by aggregating individuals' choices, such evaluation can hide potentially unrealistic disaggregate behaviours. More generally, activity schedules are high-dimensional objects. Therefore more detailed analysis is needed to validate the realism of individual behaviours.

In the above works, individual schedule quality is either omitted or evaluated through visual inspection. Our work builds on the above by implementing a more comprehensive framework of distributions for evaluation, as well as proposing additional metrics considering individual schedule quality.

\subsection{Deep Generative Modelling}

We define a generative model as a model that learns an approximation of a real data distribution, based on some observed sample of data from that distribution. A generative model can then be used to generate new, previously unobserved, samples by sampling from the modelled distribution.

Deep Generative Models (DGMs) make use of large training datasets and deep machine learning architectures to learn the distributions and joint distributions of complex high-dimensional data such as images, video, text and speech. Their depth allows for the learning of complex relationships, potentially learning excellent approximations of the real generation process. In this work, we demonstrate the application of DGMs for generating large samples of human activity schedules. There are many types of DGMs with various trade-offs briefly described below.

Auto-regressive models, such as PixelCNN by \cite{PixelCNN}, assume some ordering to predictions such that there is conditionality to previous estimates. Recurrent Neural Networks (RNNs), used to make sequential classifications, can be used as auto-regressive generative models, where at each step the model makes estimates given previously sampled estimates. This approach is popular in language modelling but assumes an order in the data generation process that may be unrealistic for scheduling.

Variational Auto Encoders (VAEs), originally proposed by \cite{vae} and \cite{rezende2014stochastic}, are a form of latent variable model. During training, the model learns a mapping from the training data to a normally distributed latent space via an encoder model as well as a mapping from this latent space to estimates of the real distribution via a decoder model. After training, new samples can then be generated by sampling from the latent space and passing through the decoder. The VAE architecture provides a flexible and expressive framework, with demonstrated results in many complex learning domains such as images and language.

The above approaches use or estimate likelihood as a training objective. Generative Adversarial Networks (GANs), originally proposed by \cite{GAN}, discard this requirement, instead using a sample-based objective using a generator and discriminator architecture. This approach is extremely expressive, allowing the generation of highly realistic data. However, GANs do not estimate density and, as such, the likelihood of generated samples is not modelled. Training GANs is also difficult, as the generator-discriminator architecture can be unstable and prone to mode collapse, where the generator produces a limited or repetitive set of outputs.

Diffusion models, such as de-noising diffusion \citep{diffusion}, split the generation process into multiple steps of noise removal. Performance is generally comparable to or better than VAEs or GANs, but more computationally expensive due to having many de-noising steps. More generally, flow-based approaches, such as normalising flows \citep{norming_flows}, seek to more efficiently generate data through sequences of transformations, but finding or learning these transformations is often difficult.

We focus on VAEs in this work due to the requirement for good density estimation, i.e. the need to produce samples of schedules that are representative in aggregate, rather than individual sample quality. This benefit of VAEs is discussed by \cite{GANdensity} and demonstrated by \cite{precandrecal}. Although VAEs are generally considered to have poorer individual sample quality than alternative approaches, we argue that this sample quality is likely adequate for schedule-generating tasks. VAEs are also preferable due to their relative speed, stability, and ease of training. In particular, they are less prone to mode collapse than GANs.

\subsection{Evaluating Deep Generative Models}
\label{sec:dgm_eval}

A common requirement of DGMs is probability density estimation, where good density estimation results in the modelled distribution being close to the distribution of the real data. Likelihood is typically used as a metric of closeness in statistical models. However, many deep learning models, including VAEs and GANs, do not have a tractable likelihood. As such, it is typically estimated from data samples. 

As per \cite{deep_eval}, estimating complex multi-dimensional probability distributions from samples is problematic. Assumptions about how probability density is estimated can bias results. More practically, reporting a single statistic for a high-dimensional generation problem limits feedback about how the model can be expected to perform or underperform in specific use cases.

DGMs are often found to exhibit a trade-off between individual sample quality and aggregate density estimation, described as precision and recall by \cite{precandrecal} and as fidelity and diversity by \cite{fidelity}. In some use cases, such as image generation, sample quality is critical, and evaluation of density estimation is limited to ensuring there is sufficient diversity in generated outputs.

More broadly, as models only approximate real distributions, it is sensible to consider how their approximations impact performance for their intended application. For this reason, we propose several interpretable distance metrics for density estimation. These can be used to make qualitative evaluations of model suitability for various applications or combined to make quantitative comparisons between models.

\subsection{Generative ML in Transport Modelling}
\label{sec:genmltransport}

Deep generative models have been applied for population synthesis tasks - the generation of joint attributes for a representative population of individuals. \cite{borysovScalablePopulationSynthesis2019} apply a VAE architecture to population synthesis. They comprehensively evaluate the density estimation of models, comparing marginal, bi-variate and tri-variate distributions. They find their approach can outperform conventional methodologies in high-dimensional cases. \cite{kimDeepGenerativeModel2023a} add to this work, also using a GAN-based approach. They formalise a feasibility-diversity trade-off, where high feasibility is the avoidance of infeasible samples, and high diversity improves the generation of missing data. They show that the model can recover missing samples.

Deep generative models have also been used to model activity schedules.
\cite{badumarfo2020differentially}  use a GAN to generate both synthetic attributes and sequences. They show promising results for marginal attribute distributions, but find that the model performs poorly at scheduling. They note that the model struggles with the complexity of learning higher-order correlations and long-range temporal dependencies.

\cite{deepactivitymodelgenerative} discretise schedules into 15-minute steps. They use a complex deep autoregressive model to generate schedules and associated activity locations conditional on a range of socio-economic attributes for members of households. They evaluate generated schedules by comparing a small set of marginal distributions and demonstrate the generated schedules in a downstream traffic simulation task. An attention-based architecture outperforms various RNN architectures, although an LSTM architecture compares similarly.

\cite{tripCGAN} use a conditional GAN architecture to predict trip purpose and socio-demographic attributes from smart card data. The use of a generative model explicitly allows for variance in the model predictions. They compare transformer architectures with temporal and spatial encodings against baseline approaches. Evaluating the model using marginal and bivariate distributions, they find that the GANs do not outperform benchmark methods. However, through further evaluation of fidelity, diversity, creativity, and accuracy, they identify some performance benefits of GANs.

\subsection{Summary of Discriminative and Generative ML Approaches for Activity Scheduling}
\label{sec:mlsummary}

Table \ref{tab:ml_schedulers} presents an overview of existing ML approaches to activity scheduling, combining both the supervised discriminative learning approaches introduced in Section~\ref{sec:mlscheduling} and the generative learning approaches in Section~\ref{sec:genmltransport}. All but one of the existing approaches are discriminative or conditional-generative, modelling part or all the variance in observed schedules using explanatory variables such as sociodemographics. They are therefore trained using pairs of attributes and schedules, or equivalent, using a supervised learning approach. Only \cite{badumarfo2020differentially} use a purely self-supervised generative approach. 

\begin{table}
    \footnotesize
    \caption{Summary of Existing Machine Learning Activity Scheduling Approaches}
    \vspace{2ex}
    \centering
    \begin{tabular}{ l l l}
        \hline
        Model Name  & Output  & Approach \\
        \hline
        \hline
        \cite{dap_ml} & activity participations & supervised discriminative \\
        \cite{trees} & activity timings & supervised discriminative \\
        \cite{koushikActivityScheduleModeling2023} & activity schedules & supervised discriminative \\
        \cite{deepactivitymodelgenerative} & activity schedules \& locations & supervised conditional-generative \\
        \cite{tripCGAN} & activity types & supervised conditional-generative \\
        \cite{badumarfo2020differentially} & activity schedules & self-supervised generative \\
        \hline
    \end{tabular}
    \label{tab:ml_schedulers}
\end{table}

We provide a purely generative approach to activity scheduling that does not require explanatory data to model variance in scheduling. We add to the generative approach by \cite{badumarfo2020differentially} by considering a VAE approach with the goal of better modelling the true distribution of real schedules, as well as novel combinations of schedule encodings and model architectures. We also introduce a comprehensive evaluation framework, designed to evaluate the suitability of generated schedules for downstream tasks.

\subsection{Generative ML for Sequence Data}

Activity schedules can be considered as sequence data, similar to text. Language models represent text as sequences of tokens representing words or parts of words, and generate new text in an auto-regressive manner. A key challenge of language models is the consideration of long-range dependencies, leading to the development of architectures suited to the retrieval of information over very long sequences of text, such as by \cite{AIAYN}.

\cite{sequenceVAE} demonstrate a sequence-generating VAE that can generate text by sampling from the latent space. They note this is effectively a combination of an auto-regressive and latent model. They show that the latent representation can learn useful information about the overall distribution of text. Variations of this architecture are used by \cite{DRAW} for generating images, and \cite{musicVAE} for generating music. In some cases, models are evaluated on the Kullback–Leibler Divergence ($D_{KL}$) of their latent layer and regeneration of test sequences; however, this is primarily used for model comparison. Ultimately, researchers employ qualitative evaluation of models based on intended application.

Activity schedules are sequences of activity types and durations, combining categorical information, as per text data, and continuous information, as per image pixel values. Schedules also typically have a fixed total duration. We build on this observation by proposing and testing a mixed-type schedule representation.

\subsection{Research Contributions}

In this paper, we contribute a novel approach to synthesising activity schedules using deep generative models. More specifically:
\begin{itemize}
    \item We demonstrate an approach to synthesising human activity schedules using VAEs that can rapidly generate realistic samples of human activity schedules.
    \item We propose an activity schedule encoding, using a continuous encoding of activity durations.
    \item  We present a comprehensive framework for evaluating the quality of model outputs using human-interpretable distance metrics such that users can reasonably assess the suitability of models for various tasks.
\end{itemize}

We additionally provide a comparison of different schedule encodings and VAE encoder/decoder architectures to assist further research into the modelling of activity schedules. Our work is reproducible using Caveat\footnote{https://github.com/big-ucl/caveat}.

\section{Methodology}
\label{sec:methodology}

\subsection{Informal Problem Definition}

We consider a sample of schedules, drawn from a population, such that it is representative. This sample would typically represent schedules from a population of individuals for a single day, such as from a structured travel demand survey. However, it could also represent any population or subset, such as multiple schedules from the same respondents or schedules for a particular demographic. For generality, we therefore do not specify the population or application within this paper.

Our goal is to use this sample to approximate the distribution such that we can draw new synthetic samples of schedules that are representative of the original population, where representativeness concerns both the realism of individual schedules and the collective distribution of all schedules in the sample.

\subsection{Activity Schedule Definition}

We define a single activity schedule~$x$ as an ordered sequence of activity types~$a_{n}$ with associated durations~$d_{n}$:

\begin{equation}
    x = [(a_{1}, d_{1}), (a_{2}, d_{2}), ... , (a_{N}, d_{N})],
\end{equation}

\noindent where $n$ indexes the position in the schedule. Activity types are limited to a finite set $\mathcal{A}$. The number of activities in the schedule (i.e. the sequence length~$N$) may vary, but the total duration should equal the time period~$T$, such that:

\begin{equation}
    \sum_{n=1}^N d_{n} = T.
\end{equation}

We use a time period~$T$ of 24 hours, starting and ending at midnight. However, the approach can be extended to different periods and durations.

\subsection{Problem Definition}
\label{sec:prob}

We consider schedules to be a random variable \( X \subset \mathcal{X} \), where \( \mathcal{X} \) is the space of all possible activity schedules. These schedules are drawn from a target distribution \( P_X \), such that we consider them to have the same distribution, which we refer to as the \emph{real distribution}. This distribution captures the process of human activity scheduling, encompassing both individual preferences and environmental constraints such as accessibility and congestion.

Let the \emph{real sample} be composed of $N_{real}$ observed schedules:
\begin{equation}
X_{\text{real}} = (x_1, x_2, \dots, x_{N_{real}}) \subset \mathcal{X}, \quad \text{with } x \sim P_X.
\end{equation}

\( X_{\text{real}} \) is used to train a generative model \( P_\theta \), parametrised by \( \theta \), which aims to approximate the real distribution \( P_X \) with a learned distribution \( \hat{P}_X \).

From the trained model, we generate a \emph{synthetic sample} of $N_{syn}$ schedules:

\begin{equation}
X_{\text{syn}} = (\hat{x}_1, \hat{x}_2, \dots, \hat{x}_{N_{syn}}), \quad \text{with } \hat{x} \sim \hat{P}_X.
\end{equation}

The modelling objective is for the learned distribution \( \hat{P}_X \) to closely approximate the real distribution \( P_X \), such that:

\begin{equation}
\hat{P}_X \approx P_X.
\end{equation}

Note that the distribution \(\hat{P}_X\) includes the density, or probability, of all possible activity schedules and the joint distributions within each schedule. As per the real sample, these are for \emph{realised} schedules, accounting for preferences and constraints. For all the following experiments, we choose the size of the synthetic sample ($N_{syn}$) to match the size of the real sample ($N_{real}$).

\subsection{Generative Modelling with Variational Auto-encoders}

\begin{figure}
    \centering
    \includegraphics[width=.6\linewidth]{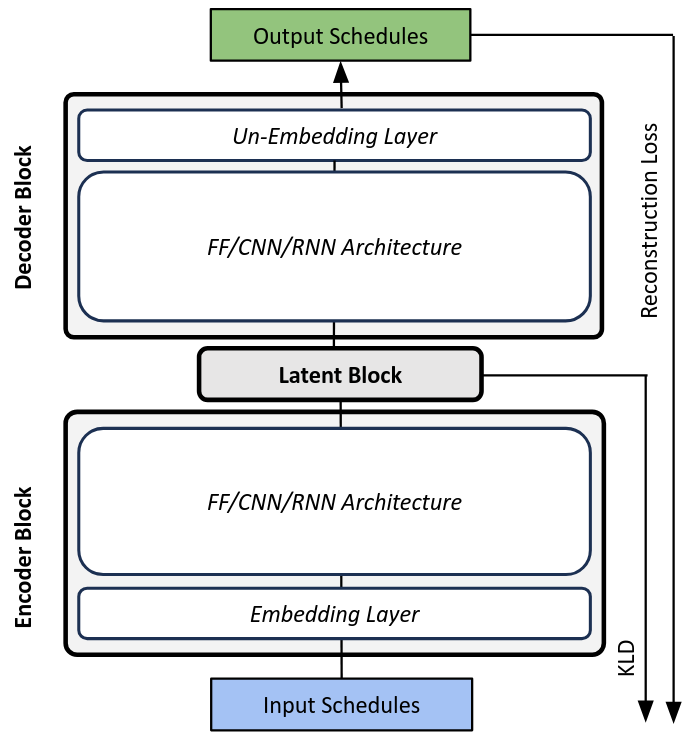}
    \caption{VAE Model Template}
    \label{fig:vae-template}
\end{figure}

We use the Variational Auto-encoder (VAE) as our deep generative modelling approach. Each VAE model is composed of an encoder block, a latent block, and a decoder block as per Figure \ref{fig:vae-template}. The various VAE encoder/decoder block architectures are detailed in Section \ref{sec:models}.

We use VAEs to generate each schedule \( x \in \mathcal{X} \) conditionally on a latent variable \( z \). The generative process is modelled as a conditional distribution \( P_\theta(X \mid Z) \), where the latent space \( Z \) is a (multivariate) standard normal, \( P(Z) = \mathcal{N}(0, I) \).

During training, an \emph{encoder block} learns a mapping from input schedules to the latent distribution. This is commonly referred to as the \emph{inference model} \( Q_\phi(Z \mid X) \). This model is responsible for mapping the real distribution of schedules to the latent space and, therefore, for learning the distribution of inter-schedule variation.

The \emph{decoder block} simultaneously learns a mapping from the latent space to output schedules. This is commonly referred to as the \emph{generative model} \( P_\theta(X \mid Z) \). This model is effectively responsible for learning the distribution of intra-schedule variation, such as temporal dependencies.

VAE encoder architectures use a stochastic process, commonly called the \emph{reparametrisation trick}, to model the random distribution $Z$. This is described in more detail in Section \ref{sec:latent_block}. Although not specifically required for the VAE approach, we use deterministic decoder architectures throughout, such that, for generation, schedules depend on sampling $Z$ only.

To learn both the inference and generative processes, VAEs use two loss functions with two aims:
\begin{inline}
    \item to reconstruct schedules passed through the entire encoder-latent-decoder structure, and
    \item to distribute the latent layer as a standard normal distribution.
\end{inline}

For the former, reconstruction loss (ReconLoss), measures the difference between each output schedule and the target input schedule. Reconstruction losses are described in Section \ref{sec:losses}. For the latter, Kullback-Leibler Divergence ($D_{KL}$) loss measures the difference between the latent layer distribution and the target normal distribution. The final loss function is composed of ReconLoss and $D_{KL}$, where $D_{KL}$ is factored using the hyperparameter~$\beta$ (reported separately for each model in Table~\ref{tab:hypers}).

\[ \text{Loss} = \text{ReconLoss} + \beta D_{KL} \]

After training, synthetic schedules are generated by first sampling \(z\), then decoding \( \hat{x} \sim P_\theta(X \mid z) \). This two-step process allows the model to capture both:
\begin{inline}
    \item \emph{Inter-schedule diversity}, through variation in the latent space \( Z \), and
    \item \emph{Intra-schedule structure}, through the conditional generation process \( P_\theta(X \mid Z) \).
\end{inline}
The sampling of \(Z\) is independent, such that there are no dependencies between schedules.

The intuition behind the VAE is that, by learning a good mapping from a known continuous latent space to the real sample of schedules. New schedule samples, with a similar distribution to the real sample, can be drawn by sampling from the latent space. For more details and a derivation of the VAE model structure and loss function, please refer to \cite{vae}.

\subsection{Observed Data}
\label{sec:data}

For the real sample of schedules, we extract and filter 59,265 24-hour trip diaries from the 2023 UK National Travel Survey (NTS) trips table\footnote{https://ukdataservice.ac.uk}. We convert this trip data into schedules using PAM \citep{PAM}. We simplify the real schedules by:
\begin{inline}
    \item  removing trips, and 
    \item simplifying activity types to the set \{home, work, education, medical, escort, other, visit, shop\}. 
\end{inline}
To remove trips, we extend the prior activity durations to include the following trip duration, such that activity start times are maintained. In effect, the following work therefore models combined activity and following trip durations. However, for conciseness, we describe this more simply as activity durations throughout. The practical implication of this is that if subsequently modelling trips, the activity duration can be treated as a budget for both the travel time and the activity participation duration.

We modify the observed data to impose structural constraints, which we later use for quantitative sample quality evaluation. The schedules are:
\begin{inline}
    \item \emph{filtered} to exclude non-home-based schedules (schedules that do not start and end at home), and 
    \item \emph{modified}, such that consecutive home, work or education activities are combined into single activities of the same type with combined duration
\end{inline}.

\begin{figure}
    \centering
    \includegraphics[trim={6.5cm 0 6.5cm 0},clip, width=1\linewidth]{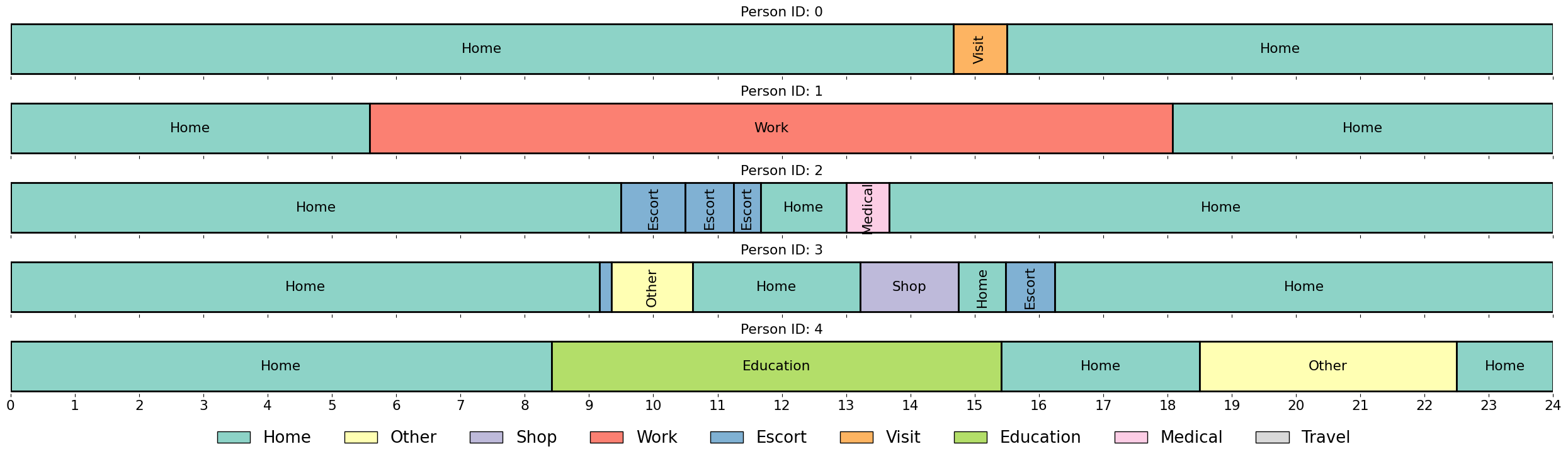}
    \caption{Example Schedules from the UK National Travel Survey (2023)}
    \label{fig:nts-examples}
\end{figure}

These constraints can be considered as \emph{structural zeros}, for example, we might expect all schedules to start and end at home and not include trips from home to home. If so, imposing them can be considered a data cleaning exercise. However, such schedules are observed, although rarely. Therefore, more pragmatically and independent of any specific broader framework requirements, these constraints can be considered as ways to evaluate model \emph{compliance} - whether the models can be constrained to structure schedules in specific ways or not. Figure ~\ref{fig:nts-examples} shows example activity sequences from the UK NTS. 

\subsection{Experiment Design}
\label{sec:experiment_design}

We train and evaluate six different VAEs with different combinations of:
\begin{inline}
    \item schedule encoding, and 
    \item model architectures.
\end{inline} These models and their naming are summarised in Table~\ref{tab:experiment}.

We consider two schedule encodings:
\begin{inline}
    \item Discrete, and 
    \item Continuous.
\end{inline} These encodings are detailed in Section \ref{sec:encodings}. We consider three model architectures:
\begin{inline}
    \item Feed-Forward (FF), composed of fully connected layers, 
    \item Convolutional (CNN), and
    \item Recurrent (RNN).
\end{inline} These architectures are detailed in Section \ref{sec:models}.

Our evaluation framework is described in Section \ref{sec:eval}. This evaluation is used for high-level comparison of schedule encodings and model architectures in Section \ref{sec:results}, and more detailed consideration of the best-performing models in Section \ref{sec:selected}. We supplement these results with different scenarios, considering data sample sizes in Section \ref{sec:size_scenarios} and alternative scenarios in Section \ref{sec:covid_scenarios}.

\begin{table}
    \footnotesize
    \caption{Summary of VAE Model Architectures}
    \vspace{2ex}
    \centering
    \begin{tabular}{ l | l | c c}
        \hline
        Model Name  & Short Name & \begin{tabular}{@{}l@{}}Schedule \\ Encoding \end{tabular} & Architecture \\
        \hline
        \hline
        Discrete FF & DiscFF & Discrete & Feed-Forward (FF) \\
        Discrete CNN & DiscCNN & Discrete & Convolutional (CNN) \\
        Discrete RNN & DiscRNN & Discrete & Recurrent (RNN) \\
        Continuous FF & ContFF & Continuous & Feed-Forward (FF) \\
        Continuous CNN & ContCNN & Continuous & Convolutional (CNN) \\
        Continuous RNN & ContRNN & Continuous & Recurrent (RNN) \\
        \hline
    \end{tabular}
    \label{tab:experiment}
\end{table}

Our experiments are intended to demonstrate:
\begin{inline}
    \item the application of a generative modelling approach to human activity scheduling, and 
    \item the impact of schedule encoding and model architecture on schedule generation quality.
\end{inline}

\section{Schedule Encodings}
\label{sec:encodings}

We consider two different schedule representations:
\begin{inline}
    \item a classic discrete encoding, and 
    \item a novel continuous encoding.
\end{inline}

\subsection{Discrete Schedule Encoding}
\label{sec:discrete}

\begin{figure}
    \centering
    \includegraphics[width=1\linewidth]{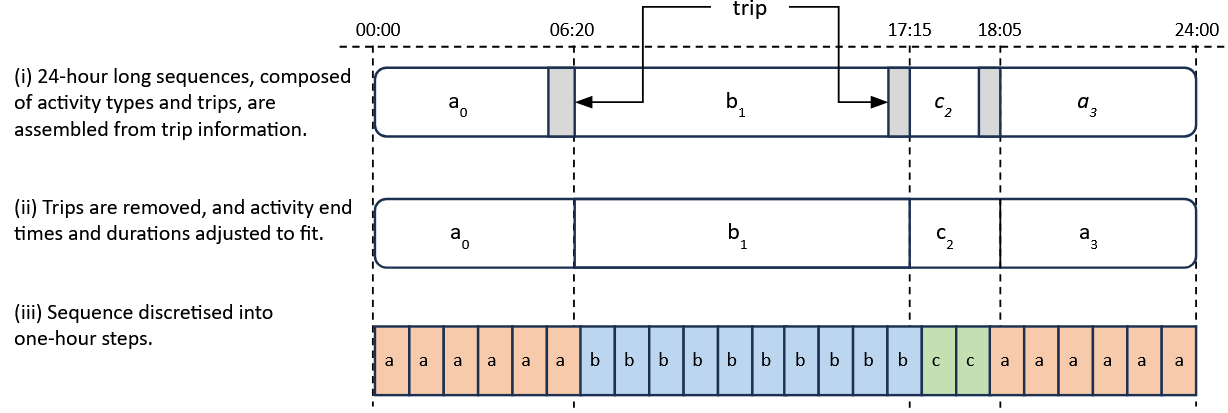}
    \caption{Discrete Activity Schedule Encoding}
    \label{fig:discrete_encoding}
\end{figure}

Individual activity schedules are typically encoded as discretised sequences, such as per \cite{koushikActivityScheduleModeling2023}. Each schedule is composed of a fixed-length vector of categorical tokens, where each token represents participating in a specific activity type for some fixed time step of duration~$t$. We use a 10-minute time step. Where a time step contains multiple activities, for example, when a schedule transitions from one activity to another, or for a very short activity, we choose the activity with the greatest duration within that time step. Figure~\ref{fig:discrete_encoding} shows an illustrative example of this encoding.

The 10-minute time step is primarily chosen to be consistent with prior work, such as by \cite{koushikActivityScheduleModeling2023}. This is additionally a pragmatic choice to allow for precise temporal predictions and the representation of short activities that might otherwise be omitted for longer step sizes. Table \ref{tab:step_eval} shows \emph{ranked} evaluations (evaluation metrics are detailed later in Section \ref{sec:eval}) for the DiscCNN model using various step sizes between 5 minutes and 2 hours. This experimentation suggests that the 10-minute step size provides a good trade-off between the evaluation metrics. 

\begin{table}
\footnotesize
    \caption{Discrete Step Size Experiment - Evaluation Rankings}
    \vspace{2ex}
    \centering
        \begin{tabular}{l | c c c c c c c }
        \hline
          & 5min & \textbf{10min} & 15min & 20min & 30min & 60min & 120min \\
        \hline
        \hline
        \multicolumn{8}{l}{}\\[-1em]
        \multicolumn{8}{l}{\textbf{Density Estimation}} \\
        \hline
        Participations & 6 & \textbf{2} & 4 & 3 & 5 & 1 & 7 \\
        Transitions & 7 & \textbf{2} & 3 & 4 & 5 & 1 & 6 \\
        Timing & 6 & \textbf{1} & 2 & 4 & 5 & 3 & 7 \\
        \hline
        \multicolumn{8}{l}{}\\[-1em]
        \textbf{Validity} & 1 & \textbf{1} & 1 & 1 & 1 & 1 & 1 \\
        \hline
        \multicolumn{8}{l}{}\\[-1em]
        \textbf{Creativity} & 2 & \textbf{1} & 3 & 4 & 5 & 6 & 7 \\
        \hline
        \end{tabular}
    \label{tab:step_eval}
\end{table}

\subsection{Continuous Schedule Encoding}
\label{sec:continuous}

The discrete schedule encoding is simple to implement but limits temporal precision to that of the time step duration~$t$. This impacts short-duration activities disproportionately, potentially removing them entirely from an encoded schedule. Reducing the step duration improves this precision but increases the encoding length. Conversely, long-duration activities are represented by many consecutive tokens representing the same activity. This is inefficient and makes correct duration prediction harder because a single wrong step cuts an activity in two, dramatically changing the schedule. We therefore propose a more precise and compact schedule encoding, representing schedules as sequences of activity types with corresponding continuous durations.

All schedule sequences start with a special \emph{start of sequence} token, followed by encodings of each activity with the associated duration. We fill the remainder of the sequence with special end-of-sequence tokens up to a maximum total sequence length. We set the maximum total sequence length to be 16, allowing for schedules with up to 15 different activities.

\begin{figure}
    \centering
    \includegraphics[width=1\linewidth]{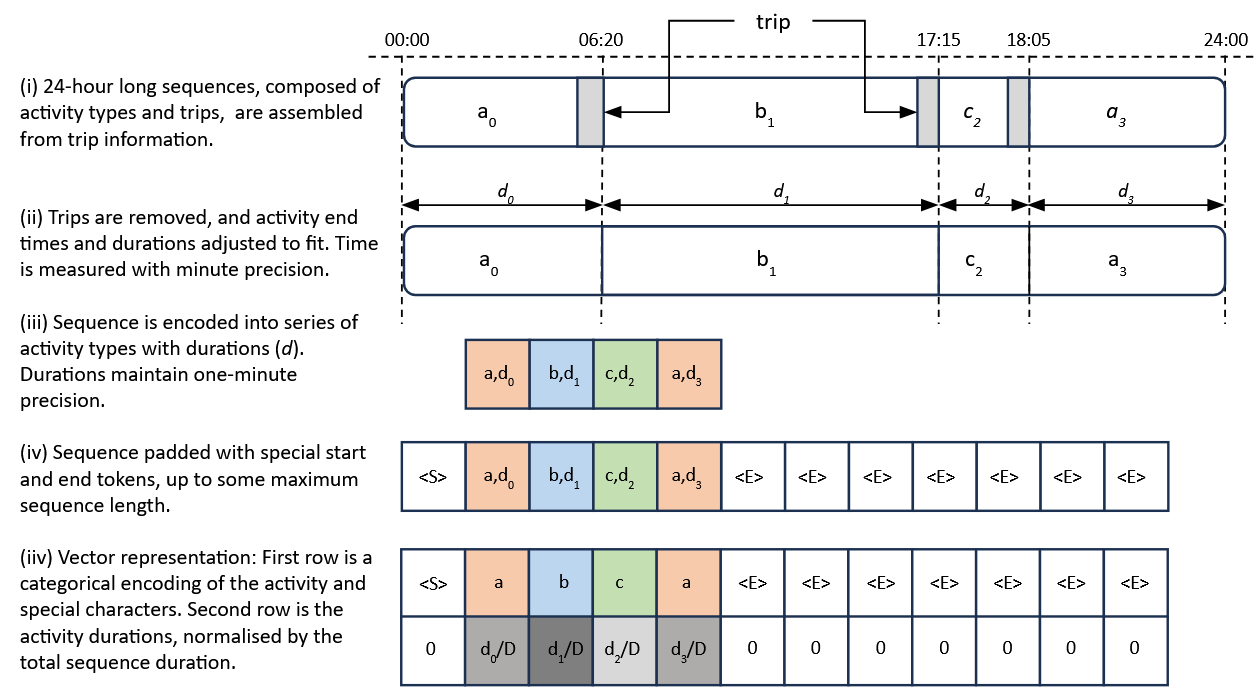}
    \caption{Sequence Activity Schedule Encoding}
    \label{fig:sequence_encoding}
\end{figure}

This approach is based on language encoding; however, it differs in that it uses a two-dimensional token. The first dimension of each token is a categorical encoding of activity type, and the second is a continuous value between 0 and 1 that represents the activity duration in days. The start and end of sequence tokens are given a duration of zero. Figure~\ref{fig:sequence_encoding} shows an illustrative example of this encoding. We refer to this as the continuous representation because it uses a continuous representation of time.

\section{Model Design}
\label{sec:models}

All models are based on a template as per Figure \ref{fig:vae-template}, composed of a latent block (Section \ref{sec:latent_block}), an encoder block and a decoder block. Depending on the schedule encoding, VAE encoder/decoder blocks use either discrete or continuous embedding and un-embedding layers as per Section \ref{sec:embed}, and alternative reconstruction losses as per \ref{sec:losses}. The various encoder/decoder block architectures are described in Section \ref{sec:architectures}. Specific language used in the following sections is defined in the mini glossary in Table \ref{tab:glos}.

\begin{table}
\footnotesize
    \caption{Model Design Glossary}
    \vspace{2ex}
    \centering
    \begin{tabular}{l l}
        \hline
         & Definition \\
        \hline
        Adam & Optimisation algorithm used for gradient descent\\
        CNN & Convolutional Neural Network \\
        Drop-out & Layer to randomly set proportion of gradients to zero during training \\
        Embedding & Layer to transform input embeddings into vector representation \\
        Feed-Forward (FF) & Fully connected layers \\
        Flatten & Operation to reduce dimensionality of input tensor \\
        $D_{KL}$ & Kullback-Leibler Divergence \\
        LeakyRELU & Activation function to add non-linearity \\
        LSTM & Long Short-Term Memory, a type of RNN \\
        Reconstruction Loss & Distance between input schedule and output schedule \\
        Resize & Layer to resize input tensor using a fully connected linear layer \\
        RNN & Recurrent Neural Network \\
        Tensor & Multi-dimensional array \\
        Un-embedding & Layer to transform input into an array of activity and duration predictions \\
        Unflatten & Operation to increase dimensionality of input tensor \\
        VAE & Variational Auto Encoder \\
        \hline
    \end{tabular}
    \label{tab:glos}
\end{table}

\subsection{Latent Block}
\label{sec:latent_block}

The VAE encoders model the latent vector $Z$ as a random variable. This cannot be directly modelled as a random sample within a VAE, because such an operation would be non-differentiable. Instead, VAEs model the latent space using learnt means and variances, allowing for differentiation; this is commonly referred to as the \emph{reparametrisation trick} and is described in more detail by \cite{vae}.

\begin{figure}
    \centering
    \includegraphics[width=.6\linewidth]{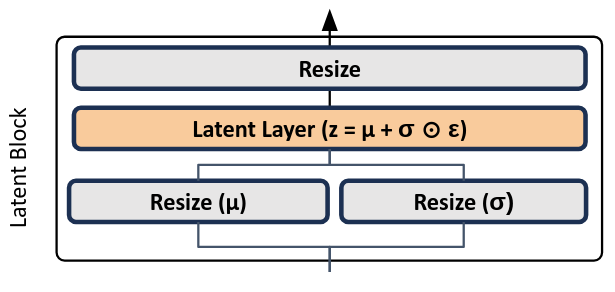}
    \caption{Latent Block Architecture}
    \label{fig:latent-template}
\end{figure}

All models use the same latent block structure shown in Figure \ref{fig:latent-template}. Within the block, the latent layer, a vector of length six, is sampled from a normal distribution with mean~$\mu$ and standard deviation~$\sigma$. $\mu$ and $\sigma$ are both vectors with the same size as the latent layer. $\mu$ and $\sigma$ are resized from the encoder output using fully connected linear layers. After sampling, the latent layer is resized as required using a fully connected linear layer for passing into the decoder block.

\subsection{Embedding and Un-Embedding}
\label{sec:embed}

The discrete and continuous schedule encodings require different embedding designs. In both cases, embedding is used to transform the input sequence encodings into two-dimensional representations. The first dimension represents the sequence length, and the second is the embedding size. In all cases, we choose the embedding size to match the layer size of the encoder/decoder architecture, reported for each model in Table \ref{tab:hypers}.

\subsubsection{Discrete Embedding and Un-Embedding}
\label{sec:discrete_embed}

Models using the discrete schedule encoding use a standard learnt embedding layer to embed each input activity token as a vector. These are then stacked to create a two-dimensional representation for each schedule.

Un-embedding reverses the operation using soft-max to output activity probabilities. When generating schedules, we sample the highest probability activity at each time step, making the decoder architecture deterministic. Note that in all cases, we find sampling the highest probabilities at each step outperforms weighted sampling.

\subsubsection{Continuous Embedding and Un-Embedding}

Models using the continuous schedule encoding use a custom embedding that embeds the first dimension (activity type token) of the input data into vectors using a learn embedding layer as above. The resulting activity embedding vectors are then concatenated back to their durations. This combines both a categorical and continuous embedding.

The continuous encoding is un-embedded by the decoder following a similar process in reverse. A vector of durations for the sequence is separated and passed through a sigmoid activation so that all duration values are between zero and one. The remaining vectors are decoded into activity probabilities using a soft-max layer.

When generating schedules, we sample the highest probability activity at each time step, making the decoder architectures deterministic. The resulting activities and their durations for each sequence are then concatenated back together. After discarding padding tokens, durations for each sequence are normalised such that they sum to one.

\subsection{Reconstruction Losses}
\label{sec:losses}

Models using the discrete embedding use activity prediction categorical cross-entropy ($CE_{acts}$) as schedule reconstruction loss. Cross-entropy is calculated between activity prediction $p(a)$ and the ground truth $q(a)$, for each activity type in $\mathcal{A}$, at each step $n$. These are then averaged across all $N$ steps. The final discrete loss, $L_{discrete}$, including $D_{KL}$ is therefore:

\begin{equation}
    L_{discrete} = CE_{acts} + \beta D_{KL}.
\end{equation}

Where:

\begin{equation}
    CE_{acts} = - \frac{1}{N} \sum_{n=1}^N \sum_{a \in \mathcal{A}}p(a_n)\log q(a_n)
\end{equation}

Models using the continuous encoding add an additional duration loss component to the reconstruction loss above. We use the mean squared error ($MSE_{durations}$) between the predicted duration $\hat{d}$ and ground truth duration $d$, for each schedule step $n$. This creates a continuous loss function, $L_{continuous}$. We introduce an additional hyperparameter $\alpha$, detailed in Table \ref{tab:hypers}, to weight the duration loss component.

\begin{equation}
L_{continuous} = CE_{acts} + \alpha MSE_{durations} + \beta D_{KL}.
\end{equation}

Where:

\begin{equation}
    MSE_{durations} = \frac{1}{N} \sum_{n=1}^N (d_n - \hat{d_n})^2
\end{equation}

\subsection{Encoder/Decoder Block Architectures}
\label{sec:architectures}

The following section summarises the Feed-Forward, Convolutional and Recurrent VAE architectures. The number of blocks ($N$) and their size ($S$) vary for each model and are summarised in Table \ref{tab:hypers}. The full architectures are available from Caveat\footnote{https://github.com/big-ucl/caveat}.

\subsubsection{Feed-Forward (FF) VAEs}
\label{sec:ff}

\begin{figure}
    \centering
    \includegraphics[width=0.6\linewidth]{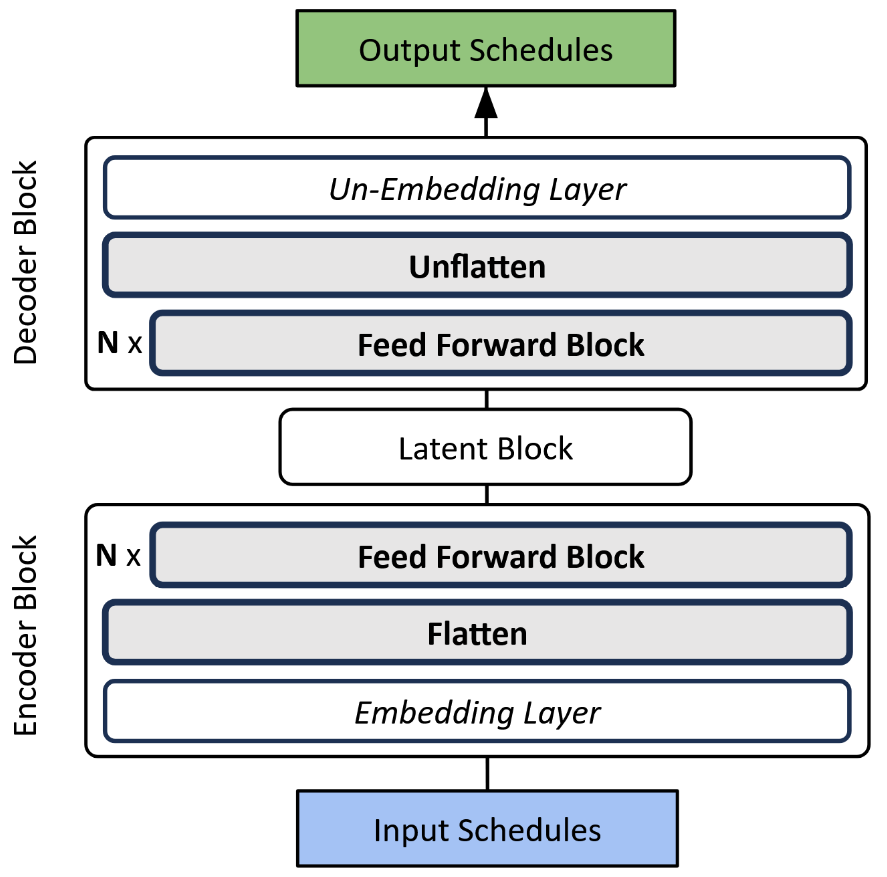}
    \caption{Feed-Forward Based VAE Template}
    \label{fig:ff-template}
\end{figure}

The Feed-Forward (FF) architecture is presented in Figure \ref{fig:ff-template}. Both encoder and decoder blocks reshape inputs or outputs as required to make use of $N$ feed-forward blocks. Each feed-forward block is composed of:
\begin{inline}
    \item a fully connected linear layer of size $S$,
    \item a batch normalisation layer,
    \item a LeakyRELU activation layer, and
    \item dropout.
\end{inline}

\subsubsection{Convolutional (CNN) VAEs}
\label{sec:cnn}

\begin{figure}
    \centering
    \includegraphics[width=0.6\linewidth]{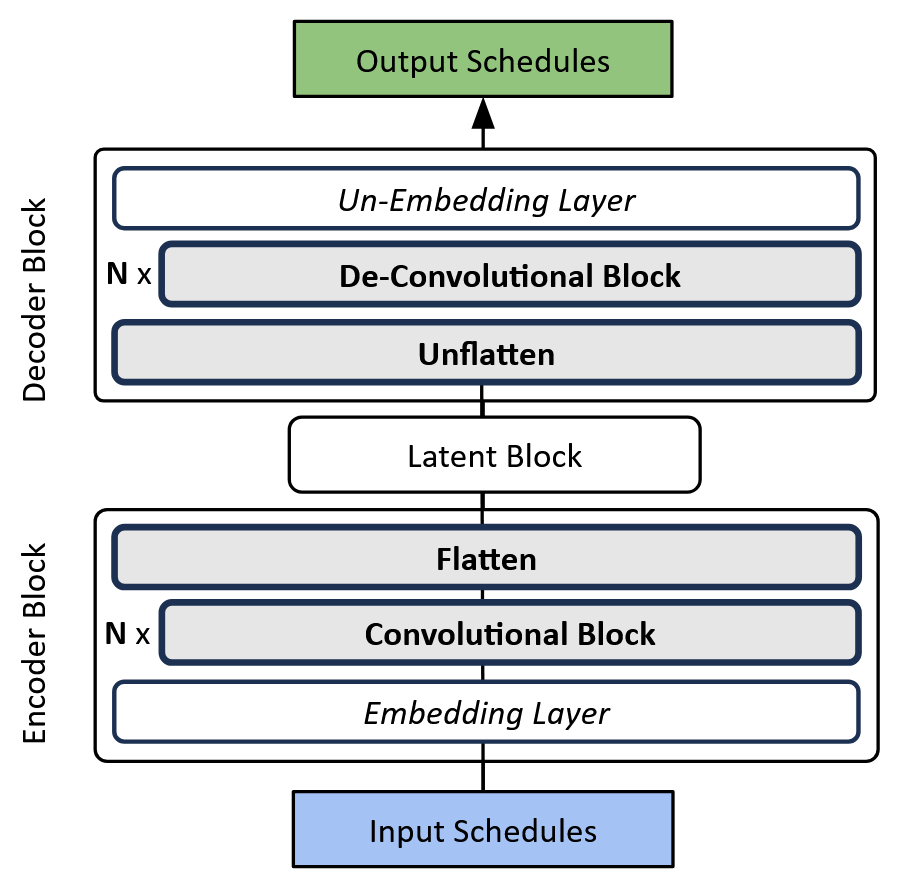}
    \caption{Convolutional Based VAE Template}
    \label{fig:cnn-template}
\end{figure}

The Convolutional (CNN) model architecture is presented in Figure~\ref{fig:cnn-template}. The encoder layer uses $N$ convolutional blocks, each composed of:
\begin{inline}
    \item a one-dimensional convolutions layer of size $S$, with stride two and convolution size four, 
    \item a batch normalisation layer,
    \item a LeakyRELU activation layer, and
    \item dropout.
\end{inline}
The two-dimensional output is then flattened for the construction of the latent block.

The decoder uses $N$ de-convolutional blocks, each as per the above convolutional blocks but using de-convolutions rather than convolutions, also of size $S$ with stride two and convolution size four. By carefully choosing output padding, the deconvolutional blocks upsample the latent layer to the required decoder output size.

\subsubsection{Recurrent (RNN) VAEs}
\label{sec:rnn}

\begin{figure}
    \centering
    \includegraphics[width=0.6\linewidth]{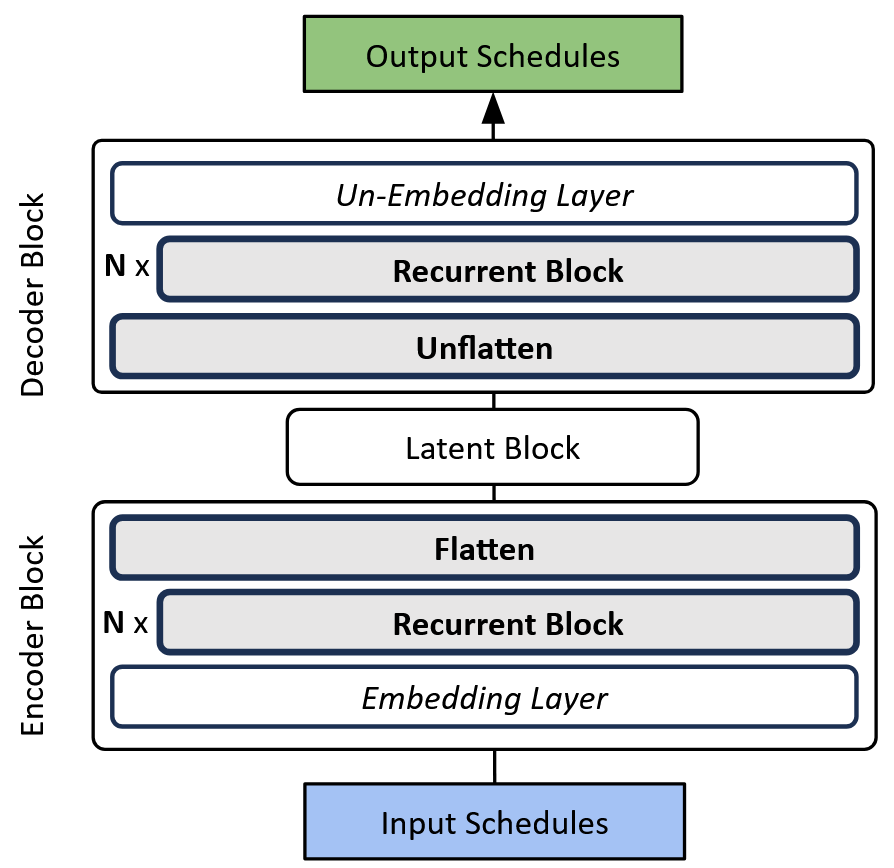}
    \caption{Recurrent Based VAE Template}
    \label{fig:rnn-template}
\end{figure}

The Recurrent (RNN) model architecture is presented in Figure ~\ref{fig:rnn-template}. The encoder layer uses $N$ stacked Long-Short Term Memory (LSTM) units of size $N$ \citep{lstm}. The block iteratively progresses through the input embeddings, and each LSTM unit passes a learnt hidden state as input to the next. The final output hidden state is then flattened as required for the latent block.

The decoder uses a similar structure in reverse. The output from the latent block is reshaped into the hidden state of the first decoding LSTM unit. This first unit is given the start of sequence token as input, further steps then use the sampled output from the previous step. During training, we use 50\% teacher forcing, where previous predictions are replaced by training data, to improve training stability.

We sample activity predictions using argmax. This approach means that the decoder block is deterministic based on input from the latent layer, so variation from the generative process only relies on sampling the VAE latent layer. Incorporating probabilistic sampling into the decoder RNNs (i.e. autoregressive generation) does not improve performance.

\subsection{Model Training and Hyperparameters}

Models are trained on 90\% of the real sample data. We use the remaining 10\% for validation during training. We train models until validation loss stabilises, typically for around 100 epochs. We do not find models suffer from over-fitting, likely due to the VAE design, which restricts information passing via the small latent layer and adds a stochastic process to training. We use Adam for gradient descent. Model training hyperparameters are reported in Table~\ref{tab:hypers}. The hyperparameters $\alpha$ and $\beta$ are chosen based on an extensive grid search to optimise VAE performances based on the evaluation criteria. The remaining hyperparameters are chosen to optimise validation loss using the Tree-structured Parzen Estimator by Optuna \citep{optuna_2019}. We therefore do not expect any significant model improvements to be possible via further hyperparameter exploration.

\begin{table}
\footnotesize
    \caption{Summary of VAE Model Hyperparameters}
    \vspace{2ex}
    \centering
    \begin{tabular}{l c c c c c c c}
        \hline
        Model & \begin{tabular}{@{}c@{}}Block \\ Size (NxS) \end{tabular}  &  \begin{tabular}{@{}c@{}}Latent \\ Size \end{tabular} & \begin{tabular}{@{}c@{}}Learning \\ Rate\end{tabular}  & \begin{tabular}{@{}c@{}}Batch \\ Size\end{tabular}  & $\beta$ & $\alpha$ & \begin{tabular}{@{}c@{}}Drop \\ Out \end{tabular} \\
        \hline
        \hline
        DicsFF & 3x32 & 6 & 0.001 & 1024 & 0.005 & - & 0.1 \\
        DiscCNN & 6x512 & 6 & 0.01 & 1024 & 0.005 & - & 0.1 \\
        DiscRNN & 4x512 & 6 & 0.001 & 1024 & 0.01 & - & 0.1 \\
        ContFF & 4x128 & 6 & 0.0001 & 1024 & 0.01 & 200 & 0.1 \\
        ContCNN & 5x128 & 6 & 0.001 & 1024 & 0.01 & 200 & 0.1 \\
        ContRNN & 4x256 & 6 & 0.001 & 1024 & 0.01 & 200 & 0.1 \\
        \hline
    \end{tabular}
    \label{tab:hypers}
\end{table}

\section{Evaluation Methodology}
\label{sec:eval}

In this section, we consider the evaluation of synthetic samples of activity schedules generated by the various models. Each model is used to generate a sample of synthetic schedules~$X_{syn}$, of equal size to the real sample~$X_{real}$.

We aim for the generated or estimated distribution of schedules in ~$X_{syn}$ to equal the distribution of schedules in $X_{real}$. This represents the ability of a model to correctly estimate the probability of any possible activity schedule - commonly called \textit{density estimation}.

As per the problem statement (Section \ref{sec:prob}), we refer to the distribution of $X_{real}$ as $P_X$, and $X_{syn}$ as $\hat{P}_X$. We estimate the similarity of $P_X$ and $\hat{P}_X$ by extracting marginal distributions from the realised samples $X_{real}$ and $X_{syn}$, then comparing these using a distance metric.

We choose marginal distributions with qualitative meaning such that we can assess the implications of the quality of density estimation on specific tasks. We consider many marginal distributions such that when distances are aggregated, we can make a reasonable quantitative evaluation of overall density estimation. Density estimation is detailed in Section \ref{sec:density}.

Perfect density estimation would satisfy our requirement for both individual schedule quality and aggregate representation across multiple schedules. However, because we can only estimate densities and because of the difficulty of directly comparing high-dimensional density estimates, we supplement evaluation by additionally considering metrics for individual schedule \emph{validity} (Section \ref{sec:sample}) and \emph{creativity} (Section \ref{sec:creativity}). This is also consistent with generative evaluation from other domains reviewed in Section \ref{sec:dgm_eval}.

\subsection{Density Estimation Evaluation}
\label{sec:density}

Our evaluation of density estimation is composed of many marginal distributions as detailed in Table \ref{tab:density}. Distributions are arranged in three domains:
\begin{itemize}
    \item \emph{Participations} which describe the participation in activity types.
    \item \emph{Transitions} which describe the order of activity type participations.
    \item \emph{Timings} which describe when activity types happen and their durations.
\end{itemize}

\subsubsection{Marginal Distributions}

Within each domain, there are multiple types of marginal distributions, each typically further segmented by activity type or types as detailed in Table \ref{tab:density}.

The participations domain considers the distributions of \begin{inline}
    \item sequence lengths - the number of activities in each schedule,
    \item \emph{single} activity participation rates - the number of times an activity of each type occurs in each schedule, 
    \item and \emph{paired} activity participation rates - the number of times pairs of activities take place in each schedule.
\end{inline} 

Sequence lengths give a quick overview of if a model is generally over or under-generating activities. Single participation rates consider the occurrences of each type of activity. We use rates rather than probability so that the occurrence of activities more than once in a schedule is also considered. For example, a schedule with a \emph{work} participation rate of two would have two distinct work activities. Paired activity rates consider the occurrences of pairs of activities within the same schedule. Note that paired activity participations do not need to be consecutive, for example, the schedule \emph{home - work - home} contains two pairs; \emph{home + work} twice and \emph{home + home} once. The consideration of non-consecutive pairs allows consideration of more complex dependencies within schedule activity participations, such as for escort and return escort trips.

The transitions domain considers schedule ordering as the distributions of sequences of activities. We consider sequences of two, three, and four consecutive activities, which we call bi, tri, and quad-grams. For example, the schedule \emph{home - work - home} contains two bi-grams; \emph{home to work} and \emph{work to home}, and a single tri-gram \emph{home to work to home}. Longer transition sequences allow consideration of longer-term dependencies within schedules.

Both participation and transition distributions are measured as \emph{rates}, which are counts of occurrences for each schedule. We then consider the distribution of such rates across the whole sample.

Timing density estimation considers the distributions of different activity type start times and durations. Note that start times and durations are segmented by activity types enumerated by position in each schedule. This allows for the sensible consideration of timing within a sequence, for example, the first home activity, \emph{home0} is distinguished from subsequent home activities, \emph{home1}, \emph{home2} and so on. We use days as a measure of time (and duration).

We additionally consider two bivariate distributions:
\begin{inline}
    \item joint start time and duration, segmented by the activity type, which we call \emph{start-durations}, and
    \item consecutive activity durations, segmented by the first activity type, which we call \emph{consecutive-durations}.
\end{inline}

Bivariate or \emph{joint} distributions allow the consideration of more complex patterns. Start-durations consider the relationships between activity start times and durations, for example, if earlier start times are typically associated with longer durations for a particular activity type. Consecutive-durations consider the relationship between an activity duration and the duration of the following activity, for example, if a short duration is typically followed by another short duration for a particular activity type.  

\begin{figure}
\footnotesize
    \centering
    \includegraphics[width=0.85\linewidth]{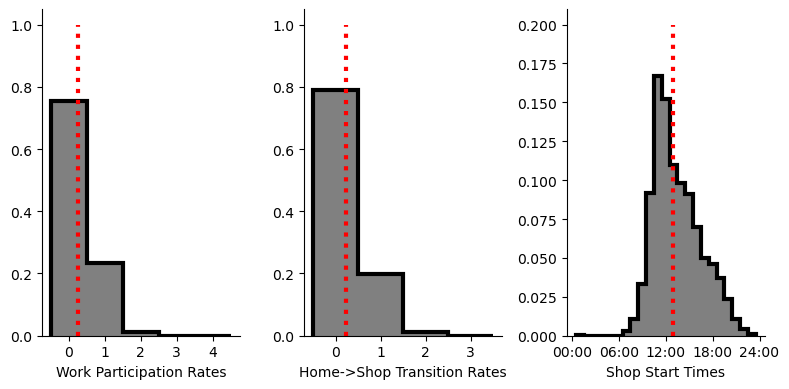}
    \caption{Example UK National Travel Survey Marginal Distributions}
    \label{fig:marginals}
\end{figure}

Figure \ref{fig:marginals} shows example marginal distributions extracted from the UK NTS. Rates are ordinal, and timings are continuous. However, the time distributions are binned at five-minute intervals to reduce computational complexity. All distributions can be described using their mean density, which can be interpreted as the expected rate or time. Expected values (mean densities) are shown in Figure \ref{fig:marginals} as vertical dashed lines.

\subsubsection{Distances}

Because it is not possible to measure the likelihood of the synthetic sample and real sample have the same distribution, we instead measure the distances between their marginal distributions as a proxy. This has the benefit of allowing for interpretable metrics, but also allows for quantitative evaluation, as lower is always better.

Distances between the real and estimated distributions are measured using Earth Movers Distance (EMD). EMD is symmetric with meaningful values, such that we can reasonably interpret them. For example, a distance of 0.1 for \emph{work} activity \emph{participation} rate can be interpreted as an expectation of 0.1 fewer or more of \emph{work} activities occurring in each schedule. Similarly, a distance of 0.5 for an activity \emph{start} can be interpreted as that activity starting, on average, half a day late or early. For bivariate distributions, we use L1 distance (also known as Manhattan\slash city-block distance).

\begin{table}
    \footnotesize
    \caption{Summary of Distributions used for Density Estimation Evaluation}
    \vspace{2ex}
    \centering
        \begin{tabular}{l | l | l | c c }
        \hline
        % Domain & \multicolumn{2}{c}{Marginal Distribution} & Units & Descriptive Metric \\
        Domain & Distribution & Segmentation & Unit & Descriptive Metric \\
        \hline
        \hline
        \rule{0pt}{1em}
        \multirow{11}{*}{\rotatebox[origin=c]{90}{\textbf{Participations}}} & Sequence Lengths & - & counts & av. length \\
                                        \cline{2-5}
                                        \rule{0pt}{1em}
                                        & \multirow{5}{*}{Single} & home & \multirow{5}{*}{rates} & \multirow{5}{*}{av. rate} \\
                                                              &  & work &  &  \\
                                                              &  & education &  &  \\
                                                              &  & shop &  & \\
                                                              &  & \rotatebox[origin=l]{90}{\textbf{...}} &  & \\
                                         \cline{2-5}
                                         \rule{0pt}{1em}
                                        & \multirow{5}{*}{Pair} & home, home & \multirow{5}{*}{rates} & \multirow{5}{*}{av. rate} \\
                                                               & & home, work &  &  \\
                                                               & & home, shop &  &  \\
                                                               & & shop, shop &  &  \\
                                                               &  & \rotatebox[origin=l]{90}{\textbf{...}} &  & \\
        \hline
        \hline
        \rule{0pt}{1em}
        \multirow{15}{*}{\rotatebox[origin=c]{90}{\textbf{Transitions}}} & \multirow{5}{*}{Bi-gram} & home--work & \multirow{5}{*}{rates} & \multirow{5}{*}{av. rate} \\
                                                      &  & work--home &  &  \\
                                                      &  & work--shop &  &  \\
                                                      &  & shop--shop &  &  \\
                                                      &  & \rotatebox[origin=l]{90}{\textbf{...}} &  & \\
                                \cline{2-5}
                                \rule{0pt}{1em}
                                & \multirow{5}{*}{Tri-gram} & home--work--home & \multirow{5}{*}{rates} & \multirow{5}{*}{av. rate} \\
                                                      &  & home--work--shop &  &  \\
                                                      &  & home--shop--work &  &  \\
                                                      &  & shop--shop--shop &  &  \\
                                                      &  & \rotatebox[origin=l]{90}{\textbf{...}} &  & \\
                                 \cline{2-5}
                                 \rule{0pt}{1em}
                                & \multirow{5}{*}{Quad-gram} & home--work--shop--home & \multirow{5}{*}{rates} & \multirow{5}{*}{av. rate} \\
                                                       & & home--shop--work--home &  &  \\
                                                       & & home--shop--shop--home &  &  \\
                                                       & & home--shop--shop--shop &  &  \\
                                                       &  & \rotatebox[origin=l]{90}{\textbf{...}} &  & \\
        \hline
        \hline
        \rule{0pt}{1em}
        \multirow{20}{*}{\rotatebox[origin=c]{90}{\textbf{Timings}}} & \multirow{5}{*}{Start times} & home0 & \multirow{5}{*}{days} & \multirow{5}{*}{av. time} \\
                                                      &  & home1 &  &  \\
                                                      &  & work0 &  &  \\
                                                      &  & \rotatebox[origin=l]{90}{\textbf{...}} &  & \\
                                \cline{2-5}
                                \rule{0pt}{1em}
                                & \multirow{5}{*}{Durations} & home0 & \multirow{5}{*}{days} & \multirow{5}{*}{av. time} \\
                                                      &  & home1 &  &  \\
                                                      &  & work0 &  &  \\
                                                      &  & \rotatebox[origin=l]{90}{\textbf{...}} &  & \\
                                 \cline{2-5}
                                 \rule{0pt}{1em}
                                & \multirow{5}{*}{Start-durations} & home & \multirow{5}{*}{days} & \multirow{5}{*}{av. time} \\
                                                      &  & work &  &  \\
                                                      &  & education &  &  \\
                                                     &  & \rotatebox[origin=l]{90}{\textbf{...}} &  & \\
                                 \cline{2-5}
                                 \rule{0pt}{1em}
                                & \multirow{5}{*}{Consecutive-durations} & home- & \multirow{5}{*}{days} & \multirow{5}{*}{av. time} \\
                                                      &  & work- &  &  \\
                                                      &  & education- &  &  \\
                                                      &  & \rotatebox[origin=l]{90}{\textbf{...}} &  & \\
        \hline
    
    \end{tabular}           
    \label{tab:density}
\end{table}

\subsubsection{Approximating Density Estimation}

Our approach of considering multiple types of marginal distributions with segmentation creates a large number of distances. Combined, these distances can give an approximation of overall density estimation. Segmented distribution distances are combined using frequency-weighted averages to provide summary metrics at the distribution level. These distribution distances can then be averaged to provide domain-level summaries. We do not further combine the three domain-level metrics as they have different units and meanings. Additionally, the domains allow qualitative consideration of model evaluations for different types of tasks, which may prioritise evaluation of one domain, such as times, over another.

\subsection{Validity Evaluation}
\label{sec:sample}

It is possible to have good density estimation but poor individual schedule quality. To evaluate this quantitatively, we impose structure on our real sample (and therefore training data) by:
\begin{inline}
    \item removing schedules that are not \emph{home-based} (i.e. that do not start and end with home activities), and
    \item combining consecutive home, work and education activities.
\end{inline}
We can then report the probability that synthetic schedules are structurally correct, based on them being both \emph{home-based} and not having consecutive home, work or education activities. We call this measure of sample quality \emph{validity} and use the probability that a schedule is \emph{invalid} as a distance metric, such that lower is better.

To account for schedules that have been previously observed by the model in the training data (and are therefore known to be structurally correct), we only report the sample quality of novel schedules - those not previously observed in the training data.

\subsection{Creativity Evaluation}
\label{sec:creativity}

We are additionally interested in the model's ability to generate both 
\begin{inline}
    \item a good \emph{diversity} of schedules -- i.e. different from one another, as well as 
    \item \emph{novel} schedules -- i.e. unique from schedules in the real sample.
\end{inline}
Diversity can be considered as a type of density evaluation metric, but is also used to specifically check if the model is suffering from mode collapse, a common issue with deep generative models, where they learn to generate an unrealistically reduced set of common outputs. We define diversity as the probability that a schedule is unique in the synthetic sample. Because diversity is not invariant to sample size, we keep all sample sizes consistent.

It is additionally desirable for our model to synthesise schedules unseen in the training data, which we define as novelty. We define novelty as the probability that a schedule is unique from the real sample. Note that the evaluation of schedule times and durations uses one-minute precision, such that synthetic samples from the discrete embedded models, which use a ten-minute precision, are structurally biased to be less creative. 

To be consistent with density estimation, where lower is better, we define homogeneity and conservatism as the opposites of diversity and novelty (as described in Table~\ref{tab:creativity}), for use as evaluation metrics, such that lower is better.

\begin{table}
\footnotesize
    \small
    \caption{Quality and Creativity Evaluation Definitions}
    \vspace{2ex}
    \centering
    \begin{tabular}{ l l }
        \hline
        Feature & Description \\
        \hline
        \hline
        Invalidity & \begin{tabular}{@{}l@{}} Probability that a schedule is not home-based or contains consecutive \\ home, work or education activities. \end{tabular}\\
        \hline
        \hline
        Homogeneity & Probability of a sequence within the synthetic sample \textbf{not being} unique. \\
        \hline
        Conservatism & Probability of a sequence \textbf{occurring} in the training sample.\\
        \hline
    \end{tabular}
    \label{tab:creativity}
\end{table}

\section{Results}
\label{sec:results}

We first summarise and compare all models at a high level. The density estimation of two selected models, the Discrete CNN and Continuous RNN, is then evaluated in more detail in Section~\ref{sec:selected}. Results are supplemented with various scenarios in Sections~\ref{sec:size_scenarios} and \ref{sec:covid_scenarios} using our preferred model, the Continuous RNN.

Table~\ref{tab:eval_summary} presents a summary of domain density estimations, sample quality and creativity for the six models. We present distances such that lower metrics are better. Models have stochastic results both from the training process and from sampling the latent space. We therefore present all evaluation descriptive and distance metrics as means from five model runs. Variance is presented in a more detailed evaluation in Section~\ref{sec:selected}.

\begin{table}
\footnotesize
    \caption{Models Evaluation Summary: Domain-level distances, such that lower is better }
    \vspace{2ex}
    \centering
        \begin{tabular}{l | c c c | c c c | l}
        \hline
          & \multicolumn{3}{ c |}{Discrete} & \multicolumn{3}{c|}{Continuous} & Unit\\
          & FF & CNN & RNN & FF & CNN & RNN &  \\   
        \hline
        \hline
        \multicolumn{8}{l}{}\\[-1em]
        \multicolumn{8}{l}{\textbf{Density Estimation}} \\
        \hline
        Participations & 0.152 & 0.162 & 0.817 & 0.062 & \textbf{0.046} & 0.064 & rate EMD \\
        & $\pm$ 0.018 & $\pm$ 0.026 & $\pm$ 0.130 & $\pm$ 0.015 & $\pm$ 0.014 & $\pm$ 0.020 & \\
        \hline
        Transitions & 0.008 & 0.006 & 0.165 & \textbf{0.004} & 0.005 & \textbf{0.004} & rate EMD \\
        & $\pm$ 0.001 & $\pm$ 0.001 & $\pm$ 0.141 & $\pm$ 0.001 & $\pm$ 0.001 & $\pm$ 0.001 & \\
        \hline
        Timing & 0.051 & 0.044 & 0.210 & 0.056 & 0.055 & \textbf{0.025} & days EMD \\
        & $\pm$ 0.008 & $\pm$ 0.005 & $\pm$ 0.054 & $\pm$ 0.001 & $\pm$ 0.001 & $\pm$ 0.009 \\   
        \hline
        \hline
        \multicolumn{8}{l}{}\\[-1em]
        \textbf{Validity} & \textbf{0.000} & \textbf{0.000} & 0.050 & 0.057 & 0.030 & 0.018 & prob. \\
        & $\pm$ 0.001 & $\pm$ 0.000 & $\pm$ 0.079 & $\pm$ 0.014 & $\pm$ 0.006 & $\pm$ 0.007 \\     
        \hline
        \hline
        \multicolumn{8}{l}{}\\[-1em]
        \textbf{Creativity} & 0.284 & 0.313 & 0.641 & 0.082 & 0.179 & \textbf{0.017} & prob. \\
        & $\pm$ 0.026 & $\pm$ 0.011 & $\pm$ 0.076 & $\pm$ 0.012 & $\pm$ 0.016 & $\pm$ 0.003 \\
        \hline
        \multicolumn{8}{l}{}\\[-1em]
        \multicolumn{8}{l}{\footnotesize(All results are means and standard deviations calculated from 5 model runs)} \\
        \end{tabular}
    \label{tab:eval_summary}
\end{table}

\subsection{Density Estimation}
\label{sec:density_estimation}

Table \ref{tab:density_dist} presents distance metrics for density estimation distributions. We see that the continuous models outperform the discrete models and that the Continuous RNN is likely preferred, as it has the lowest distances throughout, other than for sequence lengths.

For participation and transition rates, EMD can be approximately interpreted as the expected difference between the real and synthesised rates within a schedule. For example, the minimum EMD of schedule length, of the Continuous CNN, is 0.1. This can be interpreted as an expected error of 0.1 too many or too few activities per schedule. Similarly, the minimum EMD of participation is 0.033. This can be interpreted as an expected error of 3.3\% too many or too few of any activity type within a schedule.

For timing, distances can be approximately interpreted as the expected difference, in days, between the real and modelled timings of activities. For example, the minimum difference in activity durations, by the Continuous RNN, is 0.16. This can be interpreted as an expected error in activity duration of 0.016 days, or 23 minutes.

\begin{table}
\footnotesize
    \caption{Models Density Estimation Evaluation: Distribution-level distances, such that lower is better}
    \vspace{2ex}
    \centering
              \begin{tabular}{l | c c c | c c c | l}
        \hline
          & \multicolumn{3}{ c |}{Discrete} & \multicolumn{3}{c|}{Continuous} & Unit\\
          & FF & CNN & RNN & FF & CNN & RNN &  \\

        \hline \hline
        \multicolumn{8}{l}{}\\[-1em]
        \multicolumn{8}{l}{\textbf{Participations}} \\
        \hline
         Lengths & 0.363 & 0.399 & 2.100 & 0.140 & \textbf{0.100} & 0.154 & rate EMD \\
         Participation & 0.079 & 0.078 & 0.301 & 0.040 & \textbf{0.033} & \textbf{0.033} & rate EMD \\
         Pair particip. & 0.014 & 0.011 & 0.052 & 0.005 & 0.005 & \textbf{0.004} & rate EMD \\
        
        \hline \hline
        \multicolumn{8}{l}{}\\[-1em]
        \multicolumn{8}{l}{\textbf{Transitions}} \\
        \hline
        2-gram & 0.020 & 0.015 & 0.145 & 0.009 & 0.010 & \textbf{0.008} & rate EMD \\
        3-gram & 0.004 & 0.003 & 0.136 & \textbf{0.002} & 0.003 & \textbf{0.002} & rate EMD \\
        4-gram & 0.002 & 0.002 & 0.213 & \textbf{0.001} & 0.002 & \textbf{0.001} & rate EMD \\
        
        \hline \hline
        \multicolumn{8}{l}{}\\[-1em]
        \multicolumn{8}{l}{\textbf{Timing}} \\
        \hline        
        Start times & 0.022 & 0.018 & 0.028 & 0.032 & 0.033 & \textbf{0.016} & days EMD \\
        Durations & 0.039 & 0.034 & 0.215 & 0.041 & 0.039 & \textbf{0.016} & days EMD \\
        Start-durations & 0.073 & 0.067 & 0.401 & 0.071 & 0.072 & \textbf{0.031} & days EMD \\
        Joint Durations & 0.070 & 0.057 & 0.196 & 0.078 & 0.075 & \textbf{0.038} & days EMD \\
        \hline
        \multicolumn{8}{l}{\footnotesize(All results are means calculated from 5 model runs)} \\
        
        \end{tabular}
    \label{tab:density_dist}
\end{table}

\begin{figure}
    \centering
    \includegraphics[width=1\linewidth, trim={.24cm .28cm 0 -.6cm}, clip]{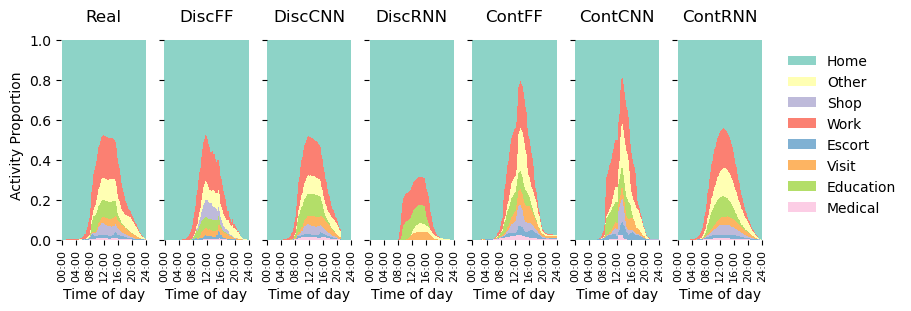}
    \caption{Comparison of Real and Synthesised Activity Frequencies: Aggregated activity participations in 10-minute time bins}
    \label{fig:freqs}
    \hspace{1cm}
    \includegraphics[width=1\linewidth, trim={.24cm .28cm 0 -.6cm}, clip]{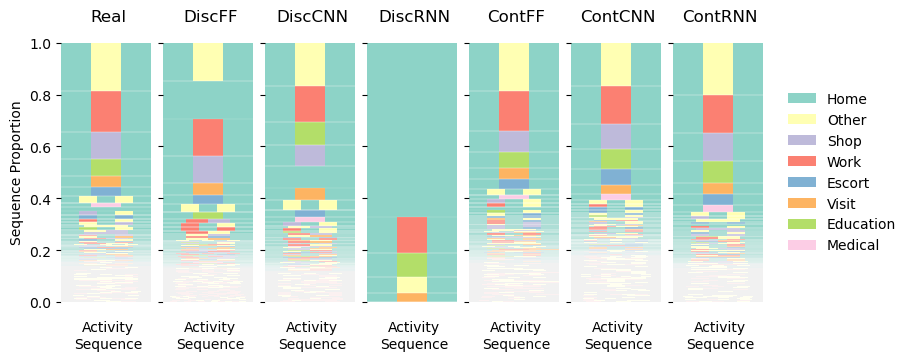}
    \caption{Comparison of Real and Synthesised Activity Sequences: Disaggregate sequences of activity types}
    \label{fig:seqs}
\end{figure}

Figures \ref{fig:freqs} and \ref{fig:seqs} present both \emph{aggregate} activity frequencies (participations binned at 10-minute intervals) and \emph{disaggregate} activity sequences, for the modelled and target real samples. Estimated distributions should look like the target real distribution. Notably, we can see that the Discrete CNN has excellent aggregate performance, but is less convincing in disaggregate, for example, introducing additional stay-at-home sequences. The introduction of such sequences is common amongst all the discrete models.

The continuous models reliably generate reasonable-looking sequences, but only the Continuous RNN, with its good timing density estimation, produces a reasonable-looking distribution of aggregate frequencies.

We speculate that the discrete models learn better aggregate patterns because time is consistently represented at each step within a sequence. This likely biases the models towards learning the most likely activity at each time step. In contrast, the continuous models cannot know the start time of each activity in a sequence without considering other predictions within the sequence. This likely biases the models towards better disaggregate sequences, but makes aggregate representation, particularly of timing, more challenging.

The DiscRNN performs the worst. This is likely because it is both 
\begin{inline}
    \item biased towards most likely activity estimation by the discrete encoding, and
    \item biased away from learning disaggregate sequence patterns due to having to pass a hidden state a relatively long way through the RNN architecture. 
\end{inline}

\subsection{Sample Quality}

Table \ref{tab:eval_quality} presents the probabilities that schedules are structurally invalid, either because they are not home-based or they contain consecutive home, work or education activities. The discrete models perform best in general. This is at least partly because they are structurally incapable of generating consecutive activities of the same type and therefore do not generate consecutive home, work or education activities. However they also no not generate non-home based schedules, with the exception of the Discrete RNN.

These results should be interpreted carefully, as the real sample has this structure artificially imposed as described in Section \ref{sec:data}. The observed NTS schedules do contain non-home-based schedules and consecutive home, work or education activities, although rarely. We can alternatively interpret the discrete models as being unable to generate consecutive activities, which could be considered negative. It is perhaps more reasonable to consider this measure of sample quality as a measure of how easily the models can be restricted to comply with some downstream requirement, such as all schedules starting and ending with home activities. In this particular case, the discrete models would be preferable.

The best performing continuous model, the Continuous RNN, produces invalid schedules only 2$\%$ of the time, and non-home-based schedules only 0.1$\%$ of the time. In a broader modelling framework, any unviable schedules would likely be either fixed or rejected and resampled, although this might introduce bias into the resulting new joint distribution. We choose not to implement such fixing or rejection sampling in this paper to better expose the trade-offs between the models.

\begin{table}
\footnotesize
    \caption{Model Validity Evaluations: Probabilities of being invalid, such that lower is better}
    \vspace{2ex}
    \centering
        \begin{tabular}{l | c c c | c c c | l}
        \hline
          & \multicolumn{3}{ c |}{Discrete} & \multicolumn{3}{c|}{Continuous} & Unit\\
          & FF & CNN & RNN & FF & CNN & RNN &  \\
        \hline
        Not home-based & \textbf{0.000} & \textbf{0.000} & 0.050 & 0.032 & 0.005 & 0.001 & prob.  \\
        Consecutive & \textbf{0.000} & \textbf{0.000} & \textbf{0.000} & 0.028 & 0.025 & 0.018 & prob. \\
        \hline
        Invalid (combined) & \textbf{0.000} & \textbf{0.000} & 0.050 & 0.057 & 0.030 & 0.018 & prob. \\  
        \hline
        \multicolumn{8}{l}{\footnotesize(All results are means calculated from 5 model runs)} \\
        \end{tabular}
    \label{tab:eval_quality}
\end{table}

\subsection{Creativity}

Table~\ref{tab:eval_creativity} presents the creativity descriptive metrics. Model creativity is considered a combination of homogeneity and conservatism. Homogeneity is the probability that generated schedules are not unique, and conservatism is the probability that generated schedules are found in the training dataset. In both cases, we consider lower as better.

The continuous models show higher creativity than the discrete models. The Continuous RNN, in particular, generates 98$\%$ unique and 99$\%$ novel schedules. However, this is partly due to the higher temporal precision of the continuous encoding.

\begin{table}
    \footnotesize
    \caption{Model Creativity Evaluations: Probabilities of being non-unique (homogeneity) or non-novel (conservatism), such that lower is better}
    \vspace{2ex}
    \centering
        \begin{tabular}{l | c c c | c c c | l}
        \hline
          & \multicolumn{3}{ c |}{Discrete} & \multicolumn{3}{c|}{Continuous} & Unit\\
          & FF & CNN & RNN & FF & CNN & RNN &  \\
        \hline
        Homogeneity & 0.468 & 0.507 & 0.931 & 0.153 & 0.348 & \textbf{0.022} & prob.  \\
        Conservatism & 0.100 & 0.118 & 0.351 & 0.011 & \textbf{0.010} & 0.012 & prob. \\
        \hline
        \multicolumn{8}{l}{\footnotesize(All results are means calculated from 5 model runs)} \\
        \end{tabular}
    \label{tab:eval_creativity}
\end{table}

\subsection{Training and Validation Losses}

Table \ref{tab:losses} shows model training and validation losses. Corresponding training and validation losses are similar for each model, showing that they are not overfit. Model reconstruction losses (CE and MSE) are fairly consistent with final evaluations. Of the discrete models, the DiscRNN underperforms the DiscFF and DiscCNN. Of the continuous models, the ContRNN outperforms the ContFF and ContCNN, but only at duration prediction, suggesting the RNN architecture provides a better structure for duration reconstruction. Note that activity losses between the discrete and continuous models are not directly comparable.

Relatively high $D_{KL}$ is generally associated with relatively lower reconstruction losses, and vice versa. This is because where models struggle with reconstruction, they can instead minimise loss by minimising the $D_{KL}$ component. For example, the DiscRNN is particularly bad at reconstruction and hence has the lowest $D_{KL}$. The continuous models partly avoid this trade-off. For example, the ContCNN has both the lowest activity loss and $D_{KL}$ (of the continuous models). This suggests that the continuous representation is more suited to this problem. Perhaps because the continuous representation of time maps better to the continuous latent space. Further work might explore models with discrete latent spaces, such as the Vector Quantised-VAE by \cite{VQVAE}.

Note that we use $\beta$ to regularise the impact of $D_{KL}$ on model learning as per Section \ref{sec:losses}. $\beta$ is carefully chosen as per Table \ref{tab:hypers} to optimise generative evaluation. During training, $\beta$ balances the tendency of the model to correctly distribute the latent layer or to reconstruct input schedules. Too low, and models excel at reconstruction, but generation evaluation is poor because the latent space is poorly distributed. Too high and models fail to reconstruct, leading to poor reconstruction losses and generation evaluation.

\begin{table}[]
    \footnotesize
    \caption{Model Training and Validation Losses Summary }
    \vspace{2ex}
    \centering
        \begin{tabular}{l | c c c | c c c }
        \hline
          & \multicolumn{3}{ c |}{Discrete} & \multicolumn{3}{c}{Continuous} \\
          & FF & CNN & RNN & FF & CNN & RNN  \\
        \hline
        \hline
        \multicolumn{7}{l}{}\\[-1em]
        \multicolumn{7}{l}{\textbf{Training Set Losses}} \\
        \multicolumn{7}{l}{}\\[-1em]
        \hline
Activity Loss (CE) & \textbf{0.106} & 0.1123 & 0.330 & 0.035 & \textbf{0.033} & 0.036 \\
 & $\pm$ 0.046 & $\pm$ 0.019 & $\pm$ 0.081 & $\pm$ 0.005 & $\pm$ 0.005 & $\pm$ 0.007 \\
Duration Loss (MSE)  & - & - & - & 0.002 & 0.002 & \textbf{0.000} \\
 & - & - & - & $\pm$ 0.000 & $\pm$ 0.000 & $\pm$ 0.000 \\
 $D_{KL}$ & 11.274 & 8.041 & \textbf{3.364} & 8.366 & \textbf{5.792} & 8.194 \\
 & $\pm$ 1.175 & $\pm$ 0.394 & $\pm$ 0.991 & $\pm$ 0.222 & $\pm$ 0.131 & $\pm$ 0.123 \\
        \hline \hline
        \multicolumn{7}{l}{}\\[-1em]
        \multicolumn{7}{l}{\textbf{Validation Set Losses}} \\
        \multicolumn{7}{l}{}\\[-1em]
        \hline
Activity Loss (CE) & \textbf{0.106} & 0.112 & 0.329 & 0.034 & \textbf{0.033} & 0.036 \\
 & $\pm$ 0.045 & $\pm$ 0.018 & $\pm$ 0.080 & $\pm$ 0.006 & $\pm$ 0.005 & $\pm$ 0.006 \\
Duration Loss (MSE)  & - & - & - & 0.002 & 0.002 & \textbf{0.000} \\
 & - & - & - & $\pm$ 0.000 & $\pm$ 0.000 & $\pm$ 0.000 \\
 $D_{KL}$ & 11.243 & 8.038 & \textbf{3.352} & 8.374 & \textbf{5.797} & 8.191 \\
 & $\pm$ 1.185 & $\pm$ 0.382 & $\pm$ 0.984 & $\pm$ 0.225 & $\pm$ 0.133 & $\pm$ 0.169 \\
        \hline
        \multicolumn{7}{l}{}\\[-1em]
        \multicolumn{7}{l}{{\footnotesize (All results are means and standard deviations from 5 model runs)}} \\
        \end{tabular}
    \label{tab:losses}
\end{table}

\section{Selected Models Density Estimation}
\label{sec:selected}

Overall, we find that the Continuous RNN is preferred for most applications. It combines both the best density estimation and the best creativity. It has only minor sample quality issues, for example, only generating 0.1$\%$ non-home-based schedules.

The following sections review the density estimation of the Continuous RNN in detail, comparing it to the best-performing discrete model, the Discrete CNN. The following tables present descriptive metrics for distributions using mean density. This is useful for qualitative evaluation but can hide differences between distributions, which are better measured using EMD or visualised in the following figures. As above, all metrics are presented as means (and standard deviations) from five model runs.

\subsection{Participations}
\label{sec:participations}

Mean densities of participation rates are compared in Table \ref{tab:participation} for the real and synthetic samples from the Discrete CNN and Continuous RNN. Participations measure the number of occurrences, i.e. the rate, of each activity type in every schedule, then use the mean density as a descriptive metric for each sample. This can be interpreted as the expected number of participations in each activity. The Continuous RNN is better than the Discrete CNN at sequence lengths and single activity participation rates. For example, the expected Continuous RNN sequence length is 3.840 activities, close to the real expected length of 3.894.

The comparison of pair participations, which consider joint dependencies within schedules (anywhere within a schedule, not necessarily consecutive), is less clear. However, as per the aggregated results in Table \ref{tab:density_dist}, the Continuous RNN is better performing overall (an expected EMD of 0.004 versus 0.011). An independent samples t-test additionally suggests a significant difference, t(5) = -8.101, p < 0.001. This is due to the Continuous RNN performing well across all activity combinations, including infrequent ones.

Note that activity pairs create a large number of combinations. For conciseness, we only show the ten most frequent and the ten least frequent. Combinations such as \emph{escort + other} occur in the synthetic samples but not in the real sample. However, we note that whilst these combinations do not occur in the real sample, there is no structural reason why they either could not or should not occur; in other words, they are \emph{sampling zeros} rather than \emph{structural zeros}. Further inspection reveals that these zero-samples are more common in the Discrete CNN sample, although they still occur infrequently.

\begin{table}
    \footnotesize
    \caption{DiscCNN and ContRNN Activity Participation Rates: Descriptions (mean densities) and Distances (EMD)}
    \vspace{2ex}
    \centering
       \begin{tabular}{l | c | ccc | ccc | l}
        \hline
          & Real & \multicolumn{3}{c |}{Discrete CNN} & \multicolumn{3}{c |}{Continuous RNN} & Unit \\
         & & mean* & std* & EMD** & mean* & std* & EMD** &  \\
        \hline \hline
        \multicolumn{9}{l}{}\\[-1em]
        \multicolumn{9}{l}{\textbf{Sequence Length}} \\
        \hline
        sequence length & 3.894 & 3.496 & 0.073 & 0.399 & 3.840 & 0.062 & \textbf{0.154} & length \\
        \hline \hline
        \multicolumn{9}{l}{}\\[-1em]
        \multicolumn{9}{l}{\textbf{Participation}} \\
        \hline
        home & 2.306 & 2.088 & 0.030 & 0.219 & 2.264 & 0.025 & \textbf{0.057} & rate \\
        other & 0.524 & 0.442 & 0.015 & 0.082 & 0.489 & 0.024 & \textbf{0.037} & rate \\
        shop & 0.291 & 0.242 & 0.041 & 0.057 & 0.298 & 0.037 & \textbf{0.036} & rate \\
        work & 0.256 & 0.295 & 0.014 & 0.040 & 0.264 & 0.019 & \textbf{0.016} & rate \\
        escort & 0.256 & 0.107 & 0.029 & 0.149 & 0.230 & 0.050 & \textbf{0.055} & rate \\
        visit & 0.128 & 0.157 & 0.026 & 0.032 & 0.146 & 0.019 & \textbf{0.031} & rate \\
        education & 0.090 & 0.114 & 0.014 & \textbf{0.026} & 0.115 & 0.008 & \textbf{0.026} & rate \\
        medical & 0.043 & 0.052 & 0.018 & 0.016 & 0.034 & 0.006 & \textbf{0.009} & rate \\
        \hline \hline
        \multicolumn{9}{l}{}\\[-1em]
        \multicolumn{9}{l}{\textbf{Pair Participation}} \\
        \hline
        home+home & 1.039 & 0.914 & 0.012 & 0.125 & 1.030 & 0.010 & \textbf{0.024} & rate \\
        other+other & 0.086 & 0.066 & 0.008 & 0.021 & 0.078 & 0.006 & \textbf{0.009} & rate \\
        home+other & 0.084 & 0.065 & 0.008 & 0.019 & 0.076 & 0.006 & \textbf{0.008} & rate \\
        escort+escort & 0.075 & 0.019 & 0.008 & 0.056 & 0.050 & 0.009 & \textbf{0.025} & rate \\
        home+escort & 0.072 & 0.019 & 0.008 & 0.054 & 0.048 & 0.008 & \textbf{0.024} & rate \\
        shop+shop & 0.033 & 0.028 & 0.012 & \textbf{0.012} & 0.047 & 0.016 & 0.014 & rate \\
        home+shop & 0.033 & 0.027 & 0.012 & \textbf{0.011} & 0.045 & 0.015 & 0.013 & rate \\
        visit+visit & 0.013 & 0.015 & 0.003 & \textbf{0.003} & 0.011 & 0.003 & \textbf{0.003} & rate \\
        home+visit & 0.013 & 0.015 & 0.003 & \textbf{0.003} & 0.011 & 0.003 & \textbf{0.003} & rate \\
        home+work & 0.011 & 0.026 & 0.005 & 0.015 & 0.014 & 0.003 & \textbf{0.003} & rate \\
       ... & ...& ...& ...& ...& ...& ...\\
        education+medical & 0.000 & 0.000 & 0.000 & \textbf{0.000} & 0.000 & 0.000 & \textbf{0.000} & rate \\
        visit+medical & 0.000 & 0.000 & 0.000 & \textbf{0.000} & 0.000 & 0.000 & \textbf{0.000} & rate \\
        education+escort & 0.000 & 0.000 & 0.000 & \textbf{0.000} & 0.000 & 0.000 & \textbf{0.000} & rate \\
        education+shop & 0.000 & 0.000 & 0.000 & \textbf{0.000} & 0.000 & 0.000 & \textbf{0.000} & rate \\
        education+work & 0.000 & 0.001 & 0.002 & 0.001 & 0.000 & 0.000 & \textbf{0.000} & rate \\
        visit+escort & 0.000 & 0.000 & 0.000 & \textbf{0.000} & 0.000 & 0.000 & \textbf{0.000} & rate \\
        medical+escort & 0.000 & 0.000 & 0.001 & \textbf{0.000} & 0.000 & 0.000 & \textbf{0.000} & rate \\
        medical+other & 0.000 & 0.000 & 0.001 & \textbf{0.000} & 0.000 & 0.000 & \textbf{0.000} & rate \\
        shop+other & 0.000 & 0.001 & 0.001 & \textbf{0.001} & 0.001 & 0.001 & \textbf{0.001} & rate \\
        escort+other & 0.000 & 0.001 & 0.002 & 0.001 & 0.000 & 0.001 & \textbf{0.000} & rate \\
        \hline
        \multicolumn{9}{l}{{\footnotesize \textasteriskcentered{ Mean and standard deviation of descriptive metric}}} \\
        \multicolumn{9}{l}{{\footnotesize \textasteriskcentered{\textasteriskcentered{ Mean distance metric}}}} \\

        \end{tabular}
    \label{tab:participation}
\end{table}

\subsection{Transitions}

Transition density estimation considers the mean densities of different activity transitions. Transitions are sequences, two, three or four activities long. Called bi-grams, tri-grams and quad-grams respectively. As per Table \ref{tab:density_dist}, the Continuous RNN outperforms the Discrete CNN at the aggregated distribution level for all cases.

Table \ref{tab:transitions} details mean densities of transition rates for the ten most and ten least frequently observed bi-grams, tri-grams and quad-grams. These represent progressively longer sequences of transitions. The mean density of a transition can be interpreted as the expected number of each transition in a schedule. For example, the most common single transition, the bi-gram \emph{home-other} has an average rate of 0.450 in the real sample. This can be interpreted as the expected number of such transitions in a schedule. The synthetic DiscCNN and ContRNN samples have expected \emph{home-other} rates of 0.414 and 0.403, underestimating the mean density of these transitions by only 0.04 of a single transition.

Similarly to participation frequencies, both models generate \emph{zero-samples} (sequences not observed in the real sample), such as \emph{work-medical-visit}. However, we also note that these do not necessarily represent \emph{structural-zeros} and could reasonably be expected to occur.

Figure \ref{fig:seqs} shows the frequencies of whole schedule activity sequences. Both the DiscCNN and ContRNN maintain the pattern of common shorter sequences and uncommon longer sequences. However, the Discrete CNN introduces a significant increase in \emph{stay-at-home} sequences, from 1.6\% in the real sample to 8.2\%. Conversely, the ContRNN under-generates \emph{stay-at-home} sequences by about 50\% (approximately 0.8\% of schedules generated by the ContRNN are \emph{stay-at-home} sequences). 

% \begin{figure}
%     \centering
%     \includegraphics[width=1\linewidth]{images/2seqs.png}
%     \caption{Whole Schedule Sequences}
%     \label{fig:2seqs}
% \end{figure}

\begin{table}
    \footnotesize
    \caption{DiscCNN and ContRNN Activity Transition Rates: Descriptions (mean densities) and Distances (EMD)}
    \vspace{2ex}
    \centering
       \begin{tabular}{l | c | ccc | ccc | l}
        \hline
          & Real & \multicolumn{3}{c |}{Discrete CNN} & \multicolumn{3}{c |}{Continuous RNN} & Unit \\
         & & mean* & std* & EMD** & mean* & std* & EMD** &  \\
        \hline \hline
        \multicolumn{9}{l}{}\\[-1em]
        \multicolumn{9}{l}{\textbf{2-grams}} \\
        \hline
        home-other & 0.450 & 0.414 & 0.010 & \textbf{0.042} & 0.403 & 0.024 & 0.048 & rate \\
        other-home & 0.444 & 0.408 & 0.011 & \textbf{0.042} & 0.394 & 0.022 & 0.050 & rate \\
        shop-home & 0.246 & 0.225 & 0.039 & 0.038 & 0.228 & 0.017 & \textbf{0.021} & rate \\
        home-work & 0.238 & 0.275 & 0.022 & 0.037 & 0.232 & 0.014 & \textbf{0.015} & rate \\
        work-home & 0.227 & 0.255 & 0.019 & 0.029 & 0.226 & 0.012 & \textbf{0.011} & rate \\
        home-shop & 0.221 & 0.190 & 0.044 & 0.049 & 0.211 & 0.019 & \textbf{0.016} & rate \\
        home-escort & 0.198 & 0.086 & 0.022 & 0.112 & 0.179 & 0.039 & \textbf{0.043} & rate \\
        escort-home & 0.190 & 0.080 & 0.022 & 0.110 & 0.159 & 0.040 & \textbf{0.051} & rate \\
        visit-home & 0.102 & 0.117 & 0.016 & \textbf{0.020} & 0.127 & 0.018 & 0.029 & rate \\
        home-visit & 0.095 & 0.101 & 0.013 & \textbf{0.012} & 0.101 & 0.013 & 0.013 & rate \\
        ... & ...& ...& ...& ...& ...& ...\\
        education-med & 0.000 & 0.001 & 0.001 & 0.001 & 0.000 & 0.000 & \textbf{0.000} & rate \\
        med-education & 0.000 & 0.001 & 0.002 & 0.001 & 0.000 & 0.000 & \textbf{0.000} & rate \\
        education-work & 0.000 & 0.017 & 0.007 & 0.017 & 0.001 & 0.001 & \textbf{0.001} & rate \\
        shop-education & 0.000 & 0.002 & 0.004 & 0.002 & 0.001 & 0.000 & \textbf{0.000} & rate \\
        work-education & 0.000 & 0.014 & 0.004 & 0.014 & 0.000 & 0.000 & \textbf{0.000} & rate \\
        \hline \hline
        \multicolumn{9}{l}{}\\[-1em]
        \multicolumn{9}{l}{\textbf{3-grams}} \\
        \hline
        home-other-home & 0.394 & 0.361 & 0.016 & \textbf{0.039} & 0.327 & 0.022 & 0.068 & rate \\
        home-work-home & 0.214 & 0.232 & 0.025 & 0.026 & 0.200 & 0.011 & \textbf{0.015} & rate \\
        home-shop-home & 0.191 & 0.163 & 0.039 & 0.044 & 0.163 & 0.021 & \textbf{0.028} & rate \\
        home-escort-home & 0.143 & 0.055 & 0.015 & 0.089 & 0.119 & 0.030 & \textbf{0.040} & rate \\
        home-education-home & 0.079 & 0.070 & 0.019 & 0.020 & 0.088 & 0.010 & \textbf{0.011} & rate \\
        home-visit-home & 0.076 & 0.076 & 0.008 & \textbf{0.007} & 0.083 & 0.010 & 0.012 & rate \\
        other-home-other & 0.051 & 0.055 & 0.009 & 0.012 & 0.040 & 0.004 & \textbf{0.011} & rate \\
        escort-home-escort & 0.039 & 0.007 & 0.004 & 0.031 & 0.025 & 0.005 & \textbf{0.013} & rate \\
        home-med-home & 0.031 & 0.029 & 0.008 & 0.006 & 0.029 & 0.004 & \textbf{0.004} & rate \\
        shop-home-other & 0.030 & 0.018 & 0.005 & 0.012 & 0.020 & 0.006 & \textbf{0.010} & rate \\
        ... & ...& ...& ...& ...& ...& ...\\
        med-visit-work & 0.000 & 0.000 & 0.000 & \textbf{0.000} & 0.000 & 0.000 & \textbf{0.000} & rate \\
        visit-escort-med & 0.000 & 0.000 & 0.000 & \textbf{0.000} & 0.000 & 0.000 & \textbf{0.000} & rate \\
        visit-med-work & 0.000 & 0.000 & 0.000 & \textbf{0.000} & 0.000 & 0.000 & \textbf{0.000} & rate \\
        visit-work-med & 0.000 & 0.000 & 0.000 & \textbf{0.000} & 0.000 & 0.000 & \textbf{0.000} & rate \\
        work-med-visit & 0.000 & 0.000 & 0.000 & \textbf{0.000} & 0.000 & 0.000 & \textbf{0.000} & rate \\
        \hline \hline
        \multicolumn{9}{l}{}\\[-1em]
        \multicolumn{9}{l}{\textbf{4-grams}} \\
        \hline
        other-home-other-home & 0.123 & 0.145 & 0.028 & \textbf{0.040} & 0.078 & 0.006 & 0.045 & rate \\
        home-other-home-other & 0.122 & 0.143 & 0.028 & 0.040 & 0.085 & 0.010 & \textbf{0.037} & rate \\
        home-escort-home-escort & 0.087 & 0.018 & 0.008 & 0.069 & 0.054 & 0.012 & \textbf{0.033} & rate \\
        escort-home-escort-home & 0.087 & 0.017 & 0.008 & 0.069 & 0.047 & 0.008 & \textbf{0.039} & rate \\
        shop-home-other-home & 0.074 & 0.046 & 0.012 & \textbf{0.028} & 0.044 & 0.017 & 0.029 & rate \\
        home-shop-home-other & 0.061 & 0.036 & 0.010 & \textbf{0.024} & 0.037 & 0.013 & \textbf{0.024} & rate \\
        home-other-home-shop & 0.051 & 0.014 & 0.005 & 0.037 & 0.038 & 0.005 & \textbf{0.013} & rate \\
        other-home-shop-home & 0.049 & 0.012 & 0.004 & 0.037 & 0.030 & 0.002 & \textbf{0.019} & rate \\
        work-home-other-home & 0.046 & 0.042 & 0.005 & \textbf{0.005} & 0.030 & 0.006 & 0.016 & rate \\
        home-work-home-other & 0.046 & 0.041 & 0.006 & \textbf{0.006} & 0.031 & 0.006 & 0.015 & rate \\
        ... & ...& ...& ...& ...& ...& ...\\
        work-visit-home-med & 0.000 & 0.000 & 0.000 & \textbf{0.000} & 0.000 & 0.000 & \textbf{0.000} & rate \\
        work-visit-med-visit & 0.000 & 0.000 & 0.000 & \textbf{0.000} & 0.000 & 0.000 & \textbf{0.000} & rate \\
        work-visit-other-med & 0.000 & 0.000 & 0.000 & \textbf{0.000} & 0.000 & 0.000 & \textbf{0.000} & rate \\
        work-visit-other-work & 0.000 & 0.000 & 0.000 & \textbf{0.000} & 0.000 & 0.000 & \textbf{0.000} & rate \\
        work-visit-work-med & 0.000 & 0.000 & 0.000 & \textbf{0.000} & 0.000 & 0.000 & \textbf{0.000} & rate \\
        \hline
        \multicolumn{9}{l}{{\footnotesize \textasteriskcentered{ Mean and standard deviation of descriptive metric}}} \\
        \multicolumn{9}{l}{{\footnotesize \textasteriskcentered{\textasteriskcentered{ Mean distance metric}}}} \\
        \end{tabular}
    \label{tab:transitions}
\end{table}

\subsection{Timing}

\begin{figure}
    \centering
    \includegraphics[width=1\linewidth]{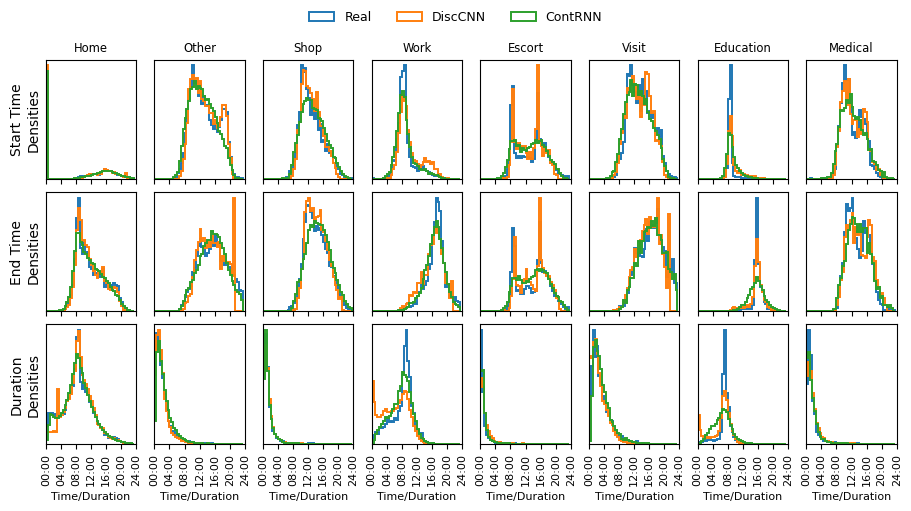}
    \caption{DiscCNN and ContRNN Activity Start Times and Durations}
    \label{fig:2times}

    \centering
    \includegraphics[width=1\linewidth]{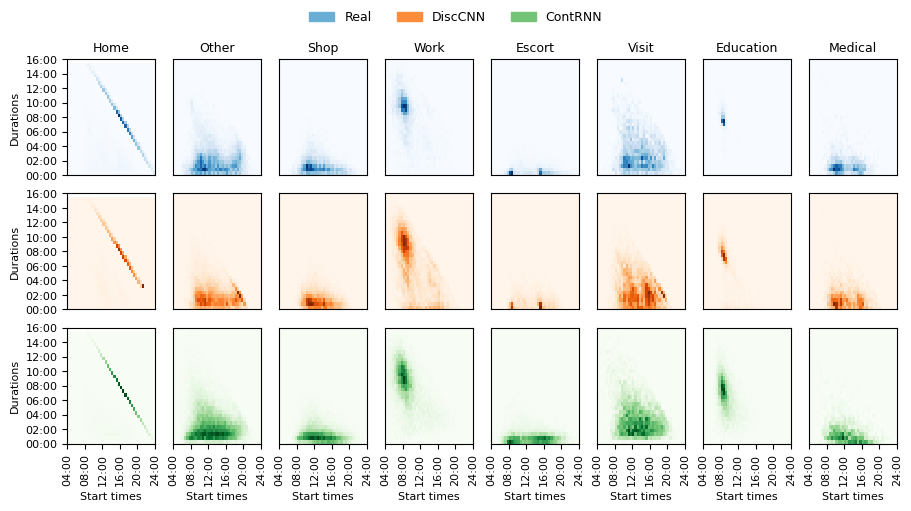}
    \caption{DiscCNN and ContRNN Joint Activity Start Times, End Times and Durations}
    \label{fig:2joint-times}
\end{figure}

Figure \ref{fig:2times} compares the distributions of start times, end times and durations for different activities. Figure \ref{fig:2joint-times} compares the joint distributions of activity start times and durations. In both cases, the synthetic samples should resemble the target real sample.

Some activities, such as \emph{shop}, \emph{education} and \emph{escort}, have more structured timing. These activities tend to take place at the same times of day, regardless of what else is in the activity sequence. The DiscCNN performs well at start and end times for these more structured activities due to its discrete representation, which ensures consistent activity timing at each step of a schedule. We consider the timing of these activities as more \emph{sequence invariant}, because they are less affected by other activities in the schedule, but more by temporal constraints such as facility opening (and closing) times, which are consistent across all schedules.

However, the DiscCNN is biased towards shorter durations, often generating unrealistic short durations for longer activities such as \emph{work} and \emph{education}. This is because long-duration activities using the discrete representation are composed of many consecutive time steps. This increases the chance of an incorrect prediction. The result of an incorrect prediction is that an activity duration is shortened into two shorter duration activities. Where activities are typically shorter, this is less of an issue, so the DiscCNN performs very well at the timing of \emph{escort} activities, which are both sequence invariant and of shorter duration.

In contrast to the DiscCNN, the ContRNN uses a continuous representation, which provides an explicit representation of activity durations, which is also directly included in the continuous loss function. This results in the ContRNN making better duration predictions than the Discrete CNN, regardless of activity length.

However, the continuous representation means that time is no longer persistent at each prediction step. Instead, it depends on both prior and later duration predictions. This makes the correct prediction of start and end times harder. Therefore, the Continuous RNN is less accurate at reconstructing activity start and end times. This is particularly notable for the more structured sequence invariant activities, such as \emph{escort}.

Table \ref{tab:times_durs} compares start times and durations for the 20 most common activities. Note that activities are enumerated, such that \emph{home0} denotes the first home activity in a sequence and \emph{home1} the second, and so on. Times (either start times or durations) are measured in days. We use the mean density as the descriptive metric, which we interpret as the expected start time or duration. For example, \emph{work0} has a real expected start time of 0.365 days (from midnight) and duration of 0.342 days, corresponding to 8:45 and just over 8 hours.

Table \ref{tab:joint-times} shows joint activity start times and durations, and joint consecutive activity durations. Consecutive durations are segmented by the type of the first activity. We use the L1 norm for these two-dimensional distributions, such that descriptions and distances can be interpreted as the sum of times and/or durations. For example, \emph{medical} has the lowest real expected joint start duration of 0.571 days. This can be interpreted as some combination of starting early and or being of short duration. These distributions also capture more subtle patterns, for example, early start times may be associated with long durations and vice versa.

As per Table \ref{tab:density_dist}, the ContRNN is better performing overall, however from Tables \ref{tab:times_durs} and \ref{tab:joint-times} we observe that the Discrete CNN tends to be better at the timing of short duration activities such as \emph{other}, \emph{shop}, \emph{visit}, and \emph{medical}, but worse at longer duration activities such as \emph{home}, \emph{work}, and \emph{education}.

\begin{table}
    \footnotesize
    \caption{DiscCNN and ContRNN Activity Start Times and Durations: Descriptions (mean densities) and Distances (EMD)}
    \vspace{2ex}
    \centering
       \begin{tabular}{l | c | ccc | ccc | l}
        \hline
          & Real & \multicolumn{3}{c |}{Discrete CNN} & \multicolumn{3}{c |}{Continuous RNN} & Unit \\
         & & mean* & std* & EMD** & mean* & std* & EMD** &  \\
        \hline \hline
        \multicolumn{9}{l}{}\\[-1em]
        \multicolumn{9}{l}{\textbf{Start Times}} \\
        \hline
        home0 & 0.000 & 0.000 & 0.000 & \textbf{0.000} & 0.000 & 0.000 & \textbf{0.000} & days \\
        home1 & 0.626 & 0.635 & 0.007 & 0.021 & 0.640 & 0.019 & \textbf{0.018} & days \\
        other0 & 0.537 & 0.539 & 0.009 & \textbf{0.014} & 0.521 & 0.006 & 0.017 & days \\
        home2 & 0.721 & 0.756 & 0.008 & 0.049 & 0.746 & 0.022 & \textbf{0.027} & days \\
        shop0 & 0.530 & 0.527 & 0.008 & \textbf{0.017} & 0.548 & 0.027 & 0.024 & days \\
        work0 & 0.365 & 0.389 & 0.007 & \textbf{0.025} & 0.388 & 0.016 & 0.028 & days \\
        escort0 & 0.499 & 0.514 & 0.013 & 0.039 & 0.505 & 0.014 & \textbf{0.020} & days \\
        visit0 & 0.562 & 0.553 & 0.004 & \textbf{0.014} & 0.573 & 0.030 & 0.023 & days \\
        education0 & 0.362 & 0.401 & 0.014 & 0.040 & 0.381 & 0.009 & \textbf{0.027} & days \\
        other1 & 0.636 & 0.668 & 0.011 & 0.035 & 0.614 & 0.019 & \textbf{0.023} & days \\
        escort1 & 0.627 & 0.619 & 0.013 & 0.036 & 0.618 & 0.012 & \textbf{0.021} & days \\
        home3 & 0.758 & 0.762 & 0.013 & \textbf{0.028} & 0.791 & 0.023 & 0.034 & days \\
        medical0 & 0.508 & 0.503 & 0.016 & \textbf{0.016} & 0.522 & 0.017 & 0.025 & days \\
        shop1 & 0.586 & 0.620 & 0.007 & 0.042 & 0.590 & 0.031 & \textbf{0.024} & days \\
        other2 & 0.673 & 0.672 & 0.015 & 0.018 & 0.664 & 0.015 & \textbf{0.017} & days \\
        escort2 & 0.662 & 0.647 & 0.026 & 0.031 & 0.662 & 0.011 & \textbf{0.014} & days \\
        visit1 & 0.641 & 0.610 & 0.020 & 0.035 & 0.649 & 0.019 & \textbf{0.021} & days \\
        work1 & 0.590 & 0.592 & 0.006 & \textbf{0.018} & 0.603 & 0.020 & 0.030 & days \\
        home4 & 0.788 & 0.754 & 0.036 & 0.046 & 0.811 & 0.021 & \textbf{0.026} & days \\
        \hline \hline
        \multicolumn{9}{l}{}\\[-1em]
        \multicolumn{9}{l}{\textbf{Durations}} \\
        \hline
        home0 & 0.456 & 0.506 & 0.007 & 0.050 & 0.457 & 0.011 & \textbf{0.013} & days \\
        home1 & 0.281 & 0.303 & 0.008 & 0.026 & 0.278 & 0.013 & \textbf{0.013} & days \\
        other0 & 0.092 & 0.092 & 0.006 & \textbf{0.008} & 0.102 & 0.010 & 0.013 & days \\
        home2 & 0.219 & 0.230 & 0.007 & 0.033 & 0.207 & 0.017 & \textbf{0.018} & days \\
        shop0 & 0.054 & 0.066 & 0.008 & 0.012 & 0.055 & 0.003 & \textbf{0.007} & days \\
        work0 & 0.342 & 0.277 & 0.006 & 0.064 & 0.318 & 0.009 & \textbf{0.027} & days \\
        escort0 & 0.035 & 0.050 & 0.005 & 0.017 & 0.042 & 0.006 & \textbf{0.013} & days \\
        visit0 & 0.128 & 0.109 & 0.004 & \textbf{0.019} & 0.157 & 0.018 & 0.029 & days \\
        education0 & 0.300 & 0.229 & 0.027 & 0.072 & 0.266 & 0.011 & \textbf{0.046} & days \\
        other1 & 0.073 & 0.069 & 0.004 & \textbf{0.008} & 0.091 & 0.013 & 0.021 & days \\
        escort1 & 0.025 & 0.044 & 0.003 & \textbf{0.019} & 0.046 & 0.002 & 0.021 & days \\
        home3 & 0.194 & 0.201 & 0.009 & 0.021 & 0.181 & 0.018 & \textbf{0.018} & days \\
        medical0 & 0.057 & 0.068 & 0.003 & \textbf{0.013} & 0.085 & 0.045 & 0.031 & days \\
        shop1 & 0.041 & 0.050 & 0.006 & 0.013 & 0.052 & 0.005 & \textbf{0.012} & days \\
        other2 & 0.059 & 0.054 & 0.006 & \textbf{0.008} & 0.079 & 0.009 & 0.022 & days \\
        escort2 & 0.025 & 0.043 & 0.007 & \textbf{0.019} & 0.044 & 0.004 & 0.020 & days \\
        visit1 & 0.091 & 0.069 & 0.005 & \textbf{0.022} & 0.119 & 0.020 & 0.031 & days \\
        work1 & 0.139 & 0.082 & 0.009 & 0.057 & 0.169 & 0.023 & \textbf{0.032} & days \\
        home4 & 0.167 & 0.200 & 0.026 & 0.044 & 0.169 & 0.014 & \textbf{0.015} & days \\
        escort3 & 0.022 & 0.040 & 0.017 & 0.023 & 0.040 & 0.004 & \textbf{0.018} & days \\
        \hline
        \multicolumn{9}{l}{{\footnotesize \textasteriskcentered{ Mean and standard deviation of descriptive metric}}} \\
        \multicolumn{9}{l}{{\footnotesize \textasteriskcentered{\textasteriskcentered{ Mean distance metric}}}} \\

        \end{tabular}
    \label{tab:times_durs}
\end{table}

\begin{table}
    \footnotesize
    \caption{DiscCNN and ContRNN Joint Activity Start Times and Durations: Descriptions (mean densities) and Distances (EMD)}
    \vspace{2ex}
    \centering
       \begin{tabular}{l | c | ccc | ccc | l}
        \hline
          & Real & \multicolumn{3}{c |}{Discrete CNN} & \multicolumn{3}{c |}{Continuous RNN} & Unit \\
         &  & mean* & std* & EMD** & mean* & std* & EMD** &  \\
        \hline \hline
        \multicolumn{9}{l}{}\\[-1em]
        \multicolumn{9}{l}{\textbf{Joint Start Durations}} \\
        \hline
        home & 0.722 & 0.741 & 0.002 & 0.074 & 0.722 & 0.006 & \textbf{0.023} & days \\
        other & 0.651 & 0.652 & 0.007 & \textbf{0.023} & 0.641 & 0.014 & 0.034 & days \\
        shop & 0.594 & 0.606 & 0.006 & \textbf{0.027} & 0.613 & 0.029 & 0.031 & days \\
        work & 0.712 & 0.670 & 0.006 & 0.115 & 0.710 & 0.011 & \textbf{0.057} & days \\
        escort & 0.584 & 0.588 & 0.010 & 0.054 & 0.584 & 0.008 & \textbf{0.033} & days \\
        visit & 0.699 & 0.667 & 0.007 & \textbf{0.036} & 0.734 & 0.040 & 0.049 & days \\
        education & 0.667 & 0.630 & 0.012 & 0.153 & 0.649 & 0.009 & \textbf{0.074} & days \\
        medical & 0.571 & 0.582 & 0.015 & \textbf{0.025} & 0.609 & 0.041 & 0.055 & days \\
        \hline \hline
        \multicolumn{9}{l}{}\\[-1em]
        \multicolumn{9}{l}{\textbf{Joint Consecutive Durations}} \\
        \hline
        home & 0.507 & 0.548 & 0.014 & 0.055 & 0.525 & 0.017 & \textbf{0.031} & days \\
        other & 0.319 & 0.335 & 0.007 & \textbf{0.028} & 0.332 & 0.016 & 0.032 & days \\
        shop & 0.330 & 0.352 & 0.020 & 0.038 & 0.305 & 0.022 & \textbf{0.037} & days \\
        work & 0.543 & 0.477 & 0.017 & 0.096 & 0.513 & 0.021 & \textbf{0.051} & days \\
        escort & 0.248 & 0.305 & 0.024 & 0.060 & 0.262 & 0.018 & \textbf{0.031} & days \\
        visit & 0.325 & 0.317 & 0.011 & \textbf{0.041} & 0.359 & 0.024 & 0.055 & days \\
        education & 0.569 & 0.425 & 0.042 & 0.150 & 0.537 & 0.013 & \textbf{0.075} & days \\
        medical & 0.331 & 0.310 & 0.040 & \textbf{0.056} & 0.421 & 0.034 & 0.098 & days \\
        \hline
        \multicolumn{9}{l}{{\footnotesize \textasteriskcentered{ Mean and standard deviation of descriptive metric}}} \\
        \multicolumn{9}{l}{{\footnotesize \textasteriskcentered{\textasteriskcentered{ Mean distance metric}}}} \\
        \end{tabular}
    \label{tab:joint-times}
\end{table}

\subsection{Density Estimation Summary}

As per Table \ref{tab:eval_summary}, we find the ContRNN to outperform the DiscRNN at all domain-level distances. We formally confirm the significance of the improvement using independent sample Welch's t-tests. We find strong evidence of improvement for participation density estimation ($p < 0.001$) and weak evidence for transitions and timing ($p < 0.1$).

\section{Scenarios and Benchmarking}

\subsection{Sample Size Scenarios}
\label{sec:size_scenarios}

To evaluate the best performing model, the Continuous RNN, with different quantities of training data, we down-sample the real population by 50\%, 25\% and 12.5\%. With the original real sample, this corresponds to real sample sizes of approximately 60,000, 30,000, 15,000 and 7,500.

For each sample size, we train and evaluate both the base Continuous RNN model and a sub-model. Each sub-model is specifically designed for each sample size. The sub-models, Continuous RNN \emph{Mid}, Continuous RNN \emph{Small} and Continuous RNN \emph{Tiny} have alternative hyperparameters specifically recalibrated for each of the sample sizes as per Table \ref{tab:sub-hypers}.

Table \ref{tab:eval_sizes_summary} summarises results from this experiment. We see a clear trend in worsening density estimation and sample quality as the sample size decreases, with a particularly steep drop-off at 12.5\%. The steep drop in performance at around 7,500 schedules likely represents a hard limit to the deep generative approach.

We show that using smaller models specifically calibrated for each sample size partly mitigates the impact of less data. By using the Continuous RNN \emph{Mid} with 30,000 schedules, performance is comparable to the full sample and base model. This is true for all evaluation metrics other than participation density estimation, which is generally improved with larger models. We expect that larger sample sizes will result in better results, although with diminishing returns.

\begin{table}
\footnotesize
    \caption{Sample Size Scenarios Hyperparameters: ContRNN Mid, Small \& Tiny}
    \vspace{2ex}
    \centering
    \begin{tabular}{l c c c c c c c}
        \hline
        Model & \begin{tabular}{@{}c@{}}Block \\ Size (NxS) \end{tabular}  &  \begin{tabular}{@{}c@{}}Latent \\ Size \end{tabular} & \begin{tabular}{@{}c@{}}Learning \\ Rate\end{tabular}  & \begin{tabular}{@{}c@{}}Batch \\ Size\end{tabular}  & $\beta$ & $\alpha$ & \begin{tabular}{@{}c@{}}Drop \\ Out \end{tabular} \\
        \hline
        \hline
        ContRNN (Base) & 4x256 & 6 & 0.001 & 1024 & 0.01 & 200 & 0.1 \\
        ContRNN Mid & 2x256 & 6 & 0.002 & 1024 & 0.005 & 200 & 0.1 \\
        ContRNN Small & 2x128 & 6 & 0.004 & 1024 & 0.0025 & 200 & 0.1 \\
        ContRNN Tiny & 2x64 & 6 & 0.008 & 1024 & 0.00125 & 200 & 0.1 \\
        \hline
    \end{tabular}
    \label{tab:sub-hypers}
\end{table}

\begin{table}
\footnotesize
    \caption{Sample Size Scenarios Evaluation Summary: Domain-level distances such that lower is better}
    \vspace{2ex}
    \centering
        \begin{tabular}{l | c | c c | c c | c c | l}
        \hline
          Model: & \multicolumn{7}{ c | }{Continuous RNN} & Unit\\
          
           & Base & Base & Mid & Base & Small & Base & Tiny & \\
          \hline
          Sample Size: & 100\% & \multicolumn{2}{ c |}{50\%} & \multicolumn{2}{ c |}{25\%} & \multicolumn{2}{ c |}{12.5\%} & \\
          
        \hline
        \hline
        \multicolumn{9}{l}{}\\[-1em]
        \multicolumn{9}{l}{\textbf{Density Estimation}} \\
        \hline
        Participations & 0.064 & \textbf{0.085} & 0.098 & \textbf{0.147} & 0.186 & \textbf{0.214} & 0.449 & rate EMD \\
        Transitions & 0.004 & 0.006 & \textbf{0.005} & 0.036 & \textbf{0.012} & 0.055 & \textbf{0.053} & rate EMD \\
        Timing & 0.025 & \textbf{0.030} & 0.035 & 0.060 & \textbf{0.057} & \textbf{0.089} & 0.121 & days EMD \\
        
        \hline
        \hline
        \multicolumn{9}{l}{}\\[-1em]
        \multicolumn{9}{l}{\textbf{Sample Quality}} \\
        \hline
        Invalid & 0.018 & 0.030 & \textbf{0.019} & 0.071 & \textbf{0.055} & 0.279 & \textbf{0.144} & prob. \\
        
        \hline
        \hline
        \multicolumn{9}{l}{}\\[-1em]
        \multicolumn{9}{l}{\textbf{Creativity}} \\
        \hline
        
        Homogenous & 0.022 & 0.009 & \textbf{0.007} & 0.086 & \textbf{0.015} & \textbf{0.070} & 0.190 & prob. \\
        Conservative & 0.012 & 0.007 & \textbf{0.006} & 0.006 & \textbf{0.003} & 0.004 & \textbf{0.003} & prob. \\
        \hline
        \end{tabular}
    \label{tab:eval_sizes_summary}
\end{table}

\begin{figure}
    \centering
    \includegraphics[width=1\linewidth, trim={.24cm .28cm 0 -.6cm}, clip]{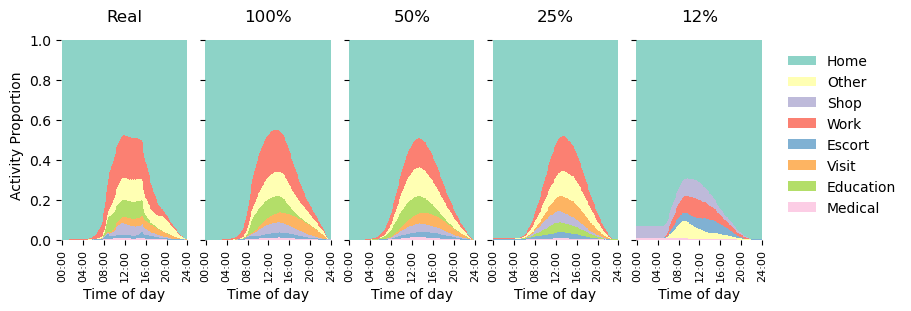}
    \caption{Sample Sizes Scenarios: Aggregate activity frequencies }
    \label{fig:sizes_freqs}
    \hspace{1cm}
    \includegraphics[width=1\linewidth, trim={.24cm .28cm 0 -.6cm}, clip]{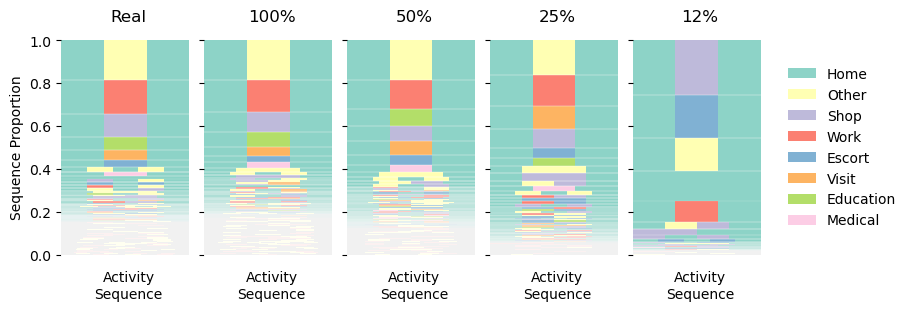}
    \caption{Sample Sizes Scenarios: Disaggregate sequences of activity types}
    \label{fig:sizes_seqs}
\end{figure}

\subsection{Alternative Year Scenarios}
\label{sec:covid_scenarios}

We evaluate how well the Continuous RNN performs on new datasets it hasn't seen before. Specifically, we train and test the model using data from 2019 and 2020. For each dataset, we retrain the model without changing any of its hyperparameters. This approach shows that the model architecture and hyperparameters are not overfit to the original data and that it can generalise effectively to new situations.

We extract schedules from the NTS travel diaries as per Section \ref{sec:data}, but create two new real samples for 2019 and 2020. We chose these samples as we expect a change in schedule distributions due to COVID-19 from 2019 to 2020. We name these samples \emph{Real 2019} and \emph{Real 2020}. Both samples have 23,452 schedules, significantly less than the base 2023 sample, so we train a Continuous RNN \emph{Mid} (Table \ref{tab:sub-hypers}) for each sample year, then generate synthetic samples from both. We name these samples \emph{Synthetic 2019} and \emph{Synthetic 2020}. 

Table \ref{tab:covid_scenario_description} summarises mean densities for the real and synthetic 2019 and 2020 scenario distributions, which can be interpreted as expected values for the various distributions shown. We consider whether the changes in schedules from 2019 and 2020 observed in the real samples are also present in the synthetic samples. To consider real and synthetic changes between 2019 and 2020, we use the difference in mean densities rather than EMD so that the direction of change can additionally be considered.

Both synthetic samples (2019 and 2020) show lower levels of participation than their respective real samples. This is consistent with previous results in Section \ref{sec:participations} and can likely be improved with more training data as per Section \ref{sec:size_scenarios}. The \emph{change} from 2019 to 2020 is well represented in the synthetic populations. Most notably, from 2019 to 2020, the real expected sequence length drops from 3.887 to 3.717 (-0.171). The synthetic samples show a similar drop from 3.700 to 3.527 (-0.173). Single and pair participations show a similar pattern, suggesting that although the models appear biased to lower activity participation, this bias is consistent across scenarios, such that the model responds to new data and therefore behaviour change is captured.

Transitions are extremely similar between the real samples, so they are not considered. For timing estimation, the models capture the real sample distributions well and therefore also show similar changes between 2019 and 2020. For example, there is a shift to activities starting 0.016 days (23 minutes) earlier in real samples, and 0.015 days (22 minutes) earlier in the synthetic samples.

This experiment suggests that the approach can capture differences between samples of schedules. Specifically, we show that the model design (both architecture and hyperparameters) has not been overfit to the original 2023 scenario and can be re-trained to capture new patterns of schedules, such as for 2019 and 2020. However, we also show that the reported differences between synthetic and real samples, particularly for participations, although small, are comparable to changes observed between 2019 and 2020 due to COVID-19.

\begin{table}
\footnotesize
    \caption{Alternative Year Scenarios: 2019 \& 2020 distribution-level descriptions (mean densities) and differences }
    \vspace{2ex}
    \centering
        \begin{tabular}{l | ccc | ccc | l}
        \hline
         & \multicolumn{3}{c |}{Real} & \multicolumn{3}{c|}{Synthetic} & Unit \\
         & 2019 & 2020 & Diff & 2019 & 2020 & Diff & \\
        \hline \hline
        \multicolumn{8}{l}{}\\[-1em]
        \multicolumn{8}{l}{\textbf{Participations}} \\
        \hline
        sequence length & 3.887 & 3.717 & -0.171 & 3.700 & 3.527 & -0.173 & rate \\
        participation rate & 0.486 & 0.465 & -0.021 & 0.462 & 0.441 & -0.022 & rate  \\
        pair participation rate & 0.041 & 0.039 & -0.002 & 0.037 & 0.036 & -0.002 & rate  \\

        \hline \hline
        \multicolumn{8}{l}{}\\[-1em]
        \multicolumn{8}{l}{\textbf{Transitions}} \\
        \hline
        
        2-gram & 0.049 & 0.046 & -0.002 & 0.049 & 0.050 & 0.001 & rate  \\
        3-gram & 0.006 & 0.006 & 0.000 & 0.007 & 0.008 & 0.001 & rate  \\
        4-gram & 0.002 & 0.003 & 0.000 & 0.003 & 0.005 & 0.001 & rate  \\

        \hline \hline
        \multicolumn{8}{l}{}\\[-1em]
        \multicolumn{8}{l}{\textbf{Timing}} \\
        \hline
        start times & 0.428 & 0.412 & -0.016 & 0.427 & 0.413 & -0.015 & days  \\
        start-durations & 0.690 & 0.686 & -0.004 & 0.700 & 0.698 & -0.001 & days \\
        durations & 0.257 & 0.269 & 0.012 & 0.270 & 0.284 & 0.013 & days \\
        joint-durations & 0.434 & 0.445 & 0.011 & 0.460 & 0.470 & 0.010 & days \\
        
        \hline

        \multicolumn{8}{l}{{\footnotesize (All results are mean descriptive metrics from 5 model runs)}} \\
        \end{tabular}
    \label{tab:covid_scenario_description}
\end{table}

% \begin{table}
%     \caption{COVID Scenario Activity Participation Rates}
%     \vspace{2ex}
%     \centering
%         \begin{tabular}{l | ccc | ccc}
%         \hline
%          & \multicolumn{3}{c |}{Real} & \multicolumn{3}{c }{Synthetic} \\
%          & 2019 & 2020 & Diff & 2019 & 2020 & Diff \\
%         \hline
%             education & 0.095 & 0.077 & -0.017 & 0.107 & 0.073 & -0.034 \\
%             escort & 0.251 & 0.217 & -0.034 & 0.166 & 0.164 & -0.002 \\
%             home & 2.294 & 2.250 & -0.044 & 2.218 & 2.176 & -0.042 \\
%             medical & 0.039 & 0.041 & 0.002 & 0.048 & 0.041 & -0.007 \\
%             other & 0.468 & 0.483 & 0.015 & 0.439 & 0.390 & -0.049 \\
%             shop & 0.293 & 0.284 & -0.008 & 0.274 & 0.275 & 0.001 \\
%             visit & 0.139 & 0.123 & -0.017 & 0.176 & 0.160 & -0.015 \\
%             work & 0.310 & 0.242 & -0.068 & 0.272 & 0.247 & -0.025 \\
%         \hline
%         \end{tabular}
%     \label{tab:covid_scenario_participations}
% \end{table}

\subsection{Comparison with Supervised Discriminative Approach}
\label{sec:discriminative}

\begin{table}
    \footnotesize
    \caption{UK National Travel Survey Label Summary}
    \vspace{2ex}
    \centering
    \begin{tabular}{ l l }
        \hline
        Attribute & Categories \\
        \hline
        Gender & \{male, female, unknown\} \\
        Age & \{0-4, 5-10, 11-15, 16-19, 20-29, 30-39, 40-49, 50-69, 70+, unknown\} \\
        Car Access & \{yes, no, unknown\} \\
        Work Status & \{employed, education, unemployed\} \\
        Income (household) & \{highest, high, medium, low, lowest, unknown\} \\
        Higher Education & \{yes, no, unknown\} \\
        License & \{yes, no, unknown\} \\
        Area & \{suburban, urban, rural, Scotland, unknown\} \\
        Ethnicity & \{white, non-white, unknown\} \\
        Household Composition & \{1A, 2A, 3A+, 1A\&1C+, 2A\&1C+, 3A+\&1C+, unknown\}* \\
        WfH & \{all, most, some, none, unknown\} \\
        Mobility Difficulty & \{none, all, walk\&bus, walk\&car, bus\&car, walk, bus, car, unknown\} \\
        Wheelchair User & \{yes, no, unknown \} \\
        Season Ticket & \{subsidised, seasonal, other, none, unknown\} \\
        Health & \{very good, good, fair, bad, very bad, unknown\} \\
        Blue Badge & \{yes, no, unknown\} \\
        \hline
        \multicolumn{2}{l}{{\footnotesize \textasteriskcentered{ A: adult, C: child}}} \\
    \end{tabular}
    \label{tab:nts-labels}
\end{table}

\begin{figure}
    \centering
    \includegraphics[width=0.85\linewidth]{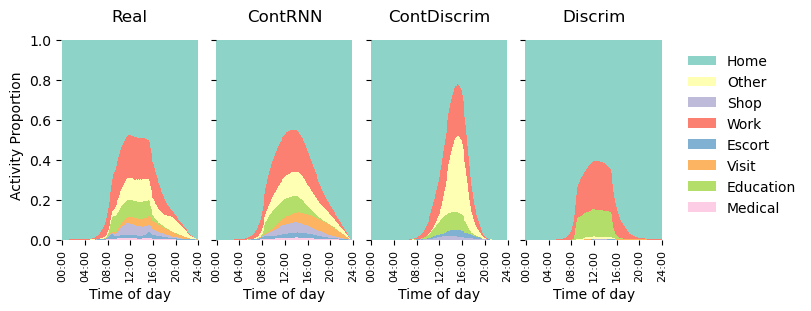}
    \caption{Comparison of Real, ContRNN and Discriminative Models: Aggregate activity frequencies}
    \label{fig:discrim_freq}
    \hspace{1cm}
    
    \includegraphics[width=0.85\linewidth]{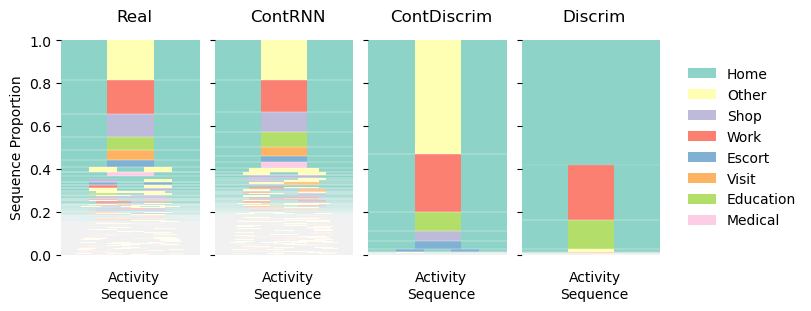}
    \caption{Comparison of Real, ContRNN and Discriminative Models: Disggregate sequences of activity types }
    \label{fig:discrim_seq}
\end{figure}

To demonstrate the advantages of our generative approach for \emph{synthesising} realistic distributions of schedules over discriminative approaches, we compare our best-performing generative model, ContRNN, against two supervised discriminative models\footnote{We note that no code has been made available for the previous ML activity scheduling approaches discussed in Section \ref{sec:mlsummary}. We have therefore replicated the methodology of \cite{koushikActivityScheduleModeling2023} who describe the experimental methodology in sufficient detail to be able to do so.}:
\begin{inline}
    \item a discriminative model developed using the methodology of \cite{koushikActivityScheduleModeling2023} with discretised time, which we call Discrim, and
    \item a second discriminative model that makes use of the continuous activity encoding, which we call ContDiscrim.
\end{inline}

Both the discriminative models predict schedules based on input sociodemographic attributes. Specifically, they seek to infer the most likely activity (and duration) at each step in a sequence, based on input attributes. Both models are trained using a standard supervised training dataset of paired attributes and schedules. To create a training dataset, we pair the NTS schedules, described in Section \ref{sec:data}, with their associated attributes, extracted from NTS individual and household data tables, and described in Table \ref{tab:nts-labels}. Attributes are treated as nominal categories, one-hot encoded and then concatenated to create an input vector for each schedule.

The Discrim model is described in detail by \cite{koushikActivityScheduleModeling2023}. Notably, schedules use the discrete encoding with 10-minute steps.  The model is composed of three bidirectional LSTM layers of size 200 and 30\% dropout. Activity predictions at each step are made as per the discrete embedding described in Section \ref{sec:discrete_embed}. The model is trained to minimise mean cross-entropy. Individual activity losses are weighted by inverse frequency in the training dataset.

The ContDiscrim model updates the design of the Discrim model to a continuous schedule encoding. This performs better than the Discrim model and enables fairer comparison to the ContRNN VAE. The model is composed of four LSTM layers of size 128 and 10\% dropout. Activity and duration predictions are made as per the continuous embedding in Section \ref{sec:continuous}. The loss function includes duration loss as per the continuous models reconstruction loss in Section \ref{sec:losses}.

For evaluation, we train the models, generate schedules, and then evaluate them using our generative evaluation methodology described in Section \ref{sec:eval}. To simulate the generative or synthesis case, the discriminative models generate schedules from the training dataset attributes. Note that in a typical supervised learning framework, we would seek to evaluate against a withheld dataset of attributes and schedules. However, in our experiment, we consider a generative or synthesis application. This also allows comparison with the generative ContRNN VAE.

\begin{table}
\footnotesize
    \caption{Comparison of ContRNN to discriminative models: domain-level distances such that lower is better}
    \vspace{2ex}
    \centering
        \begin{tabular}{l | c | c c | l}
        \hline
        & Generative & \multicolumn{2}{c|}{Discriminative} & \\
            & ContRNN & ContDiscrim & Discrim & Unit\\          
        \hline
        \hline
        \multicolumn{5}{l}{}\\[-1em]
        \multicolumn{5}{l}{\textbf{Density Estimation}} \\
        \hline
        Participations & \textbf{0.062} & 0.377 & 0.796 & rate EMD \\
        Transitions & \textbf{0.004} & 0.018 & 0.026 & rate EMD \\
        Timing & \textbf{0.024} & 0.152 & 0.196 & days EMD \\
        \hline
        \hline
        \multicolumn{5}{l}{}\\[-1em]
        \multicolumn{5}{l}{\textbf{Sample Quality}} \\
        \hline
        Invalid & 0.019 & \textbf{0.002} & 0.004 & prob. \\
        \hline
        \hline
        \multicolumn{5}{l}{}\\[-1em]
        \multicolumn{5}{l}{\textbf{Creativity}} \\
        \hline
        Homogenous & \textbf{0.022} & 0.879 & 0.975 & prob. \\
        Conservative & \textbf{0.013} & 0.016 & 0.402 & prob. \\
        \hline
        \multicolumn{5}{l}{{\footnotesize (All results are mean distances from 5 model runs)}} \\
        \end{tabular}
    \label{tab:eval_discim_summary}
\end{table}

Figures \ref{fig:discrim_freq} and \ref{fig:discrim_seq} show that both discriminative models fail to create a realistic distribution of schedules. This is confirmed by domain-level density evaluation in Table \ref{tab:eval_discim_summary}. We can explain the poor performance of the Discrim model as a combination of the discrete embedding and relatively long RNN, which bias the model towards predicting the most likely activity at each time step, which is \emph{home}. This is discussed in more detail in Section \ref{sec:density_estimation}.

The ContDiscrim model also performs worse than the ContRNN, despite being architecturally comparable, suggesting that without the explicit incorporation of a generative capability (through the latent space in the VAE), the models struggle to learn a realistic distribution of schedules.

\section{Results Discussion}
\label{sec:practical}

\begin{table}
\footnotesize
    \caption{Model Sizes and Run Times for different generative model architectures}
    \vspace{2ex}
    \centering
    \begin{tabular}{l c c c c}
        \hline
        Model & \begin{tabular}{@{}c@{}}Trainable Params \\ (millions) \end{tabular}  &  \begin{tabular}{@{}c@{}}Memory \\ (Mb) \end{tabular} & \begin{tabular}{@{}c@{}}Training \\ Time (min)\end{tabular}  & \begin{tabular}{@{}c@{}}Generation \\ Time (sec)\end{tabular} \\
        \hline
        \hline
        DicsFF & 116.2 & 467.8 & 4.3 & 1.4 \\
        DiscCNN & 8.5 & 33.8 & 4.8 & 1.2 \\
        DiscRNN & 16.9 & 67.6 & 95.8 & 8.9 \\
        ContFF & 30.2 & 120.7 & 2.1 & 0.7 \\
        ContCNN & 0.2 & 1.0 & 2.4 & 0.5 \\
        ContRNN & 4.3 & 17.0 & 4.2 & 1.4 \\
        \hline
    \end{tabular}
    \label{tab:speed}
\end{table}

We find that encoding choice, i.e. discrete or continuous, dominates model selection. We strongly recommend the use of the continuous encoding in most cases. If a discrete encoding is being used, we suggest the use of a CNN-based architecture as per the Discrete CNN.

The best-performing model, the Continuous RNN, is composed of both the continuous schedule encoding and an RNN architecture, both of which introduce additional complexity compared to the Discrete CNN. This complexity is broadly due to treating scheduling as both a categorical problem (for activity types) and a continuous problem (for activity durations). Specifically, the continuous encoding requires an additional hyperparameter $\alpha$. However, we find that all the models are reasonably insensitive to hyperparameter selection, such that models can be quickly optimised with grid search or similar.

Hyperparameter tuning is also facilitated by fast model training times. Table \ref{tab:speed} shows model sizes and expected run times based on an NVIDIA RTX A5000 GPU. The Continuous RNN is trained in only 4.2 minutes and generates approximately 60,000 schedules in only 1.4 seconds. A full synthetic sample for the UK (68.3 million) could be generated in less than 30 seconds. This speed makes model and hyperparameter search cheap, moving the bottleneck in model development to the user and the evaluation tools. We provide Caveat\footnote{https://github.com/big-ucl/caveat} to allow reproduction of results and enable extensive model and hyperparameter exploration.

Experiments with smaller data samples in Section \ref{sec:size_scenarios} show that more data leads to better results. Smaller models can be used for smaller datasets, but the deep generative approach becomes unviable at around 7,500 schedules or fewer. Our scenarios in Section \ref{sec:covid_scenarios} show that the Continuous RNN model architecture and hyperparameters can reasonably be expected to perform well on datasets in future years.

Section \ref{sec:discriminative} shows that the generative approach is more capable of modelling the variance of real schedules than existing discriminative approaches. This is unsurprising given the complexity of the real scheduling process and the relatively limited availability of explanatory variables.  We effectively demonstrate that variance in activity schedules cannot be well modelled using available attributes alone, that discriminative models would likely benefit from incorporating generative mechanisms, and that the ContRNN should likely be favoured over discriminative models when seeking to synthesise a realistic distribution of schedules from data.

\section{Conclusions}

We demonstrate the synthesis of large, diverse samples of realistic activity schedules with a deep generative approach based on a VAE architecture. Our work demonstrates a deep generative ML model approximating a human behavioural system, with complex dependencies and constraints. Our generative approach makes rapid and diverse synthesis, upsampling or anonymisation of observed activity schedules feasible. Such samples can then be used for applied models and simulations in the transport, energy and epidemiology domains.

We use a comprehensive evaluation framework to compare alternative schedule encodings and model designs. We show that a novel continuous schedule encoding is preferable to a standard discrete encoding, although it adds complexity to model design and specification.

Our best-performing Continuous RNN model achieves both aggregate and disaggregate realism. This model combines continuous schedule encoding and a recurrent structure within the VAE encoder and decoder architectures. However, this model exhibits some poor sample quality, for example, generating non-home-based schedules 0.1\% of the time.

Depending on the application, structural zero samples might need to be removed from synthetic samples and new schedules generated, or otherwise fixed, which might introduce bias. More generally, compared to traditional approaches, which are easier to constrain, this inability to impose specific structures on generated schedules should be considered a disadvantage. However, in many cases, it is likely acceptable given the gain in speed, simplicity and realistic distribution. The choice of a discrete encoded model, which has some imposed structure, could also be considered for some specific applications.

All models can generate millions of sequences in seconds. They have minor stochastics and are not overly sensitive to hyperparameters. This makes them a practical choice for human activity modelling frameworks using synthetic schedules.

\subsection{Further Research}

For this work, we focus on a deep generative VAE architecture, which we find to have adequate capacity for this task. We note that alternative generative approaches are available and could be formally compared.

We find some trade-offs between the Discrete CNN and Continuous RNN models for the generation of times for short versus long activities. This might be purely the result of biases within the chosen encodings and architectures, but might also reflect different choice-making processes for different activities. Further work should look to remove this trade-off. For example, by adding time of day context to the Continuous RNN architecture.

The application of the current work is limited to where activity schedule generation is not conditional. For traditional transport modelling frameworks, non-conditioned generation is only adequate if schedules are assumed static. This is sometimes the case when developing short-term scenarios or if the model's focus is on subsequent transport supply simulation scenarios. However, in most transport demand modelling frameworks, conditionality is required, typically for changing socio-demographics and accessibility. Further work will therefore look to add conditionality to the generative process.

Further work should also look to increase the complexity of schedule generation. For example, extending to multiple days, incorporating additional activity types, trips, and additional attributes such as location or travel mode choice. We share Caveat\footnote{https://github.com/big-ucl/caveat} to provide a framework for further model development and evaluation.

\section{Acknowledgements}

This work was supported by the Engineering and Physical Sciences Research Council (EPSRC) [grant numbers EP/T517793/1 and EP/W524335/1].

% \appendix

% \section{Sample Appendix Section}
% \label{sec:sample:appendix}
% Lorem ipsum dolor sit amet, consectetur adipiscing elit, sed do eiusmod tempor section~\ref{sec:sample1} incididunt ut labore et dolore magna aliqua. Ut enim ad minim veniam, quis nostrud exercitation ullamco laboris nisi ut aliquip ex ea commodo consequat. Duis aute irure dolor in reprehenderit in voluptate velit esse cillum dolore eu fugiat nulla pariatur. Excepteur sint occaecat cupidatat non proident, sunt in culpa qui officia deserunt mollit anim id est laborum.

 % \bibliographystyle{elsarticle-num}
 \bibliographystyle{elsarticle-harv}
 \bibliography{cas-refs}

\begin{thebibliography}{39}
\expandafter\ifx\csname natexlab\endcsname\relax\def\natexlab#1{#1}\fi
\providecommand{\url}[1]{\texttt{#1}}
\providecommand{\href}[2]{#2}
\providecommand{\path}[1]{#1}
\providecommand{\DOIprefix}{doi:}
\providecommand{\ArXivprefix}{arXiv:}
\providecommand{\URLprefix}{URL: }
\providecommand{\Pubmedprefix}{pmid:}
\providecommand{\doi}[1]{\href{http://dx.doi.org/#1}{\path{#1}}}
\providecommand{\Pubmed}[1]{\href{pmid:#1}{\path{#1}}}
\providecommand{\bibinfo}[2]{#2}
\ifx\xfnm\relax \def\xfnm[#1]{\unskip,\space#1}\fi
%Type = Inproceedings
\bibitem[{Akiba et~al.(2019)Akiba, Sano, Yanase, Ohta and Koyama}]{optuna_2019}
\bibinfo{author}{Akiba, T.}, \bibinfo{author}{Sano, S.}, \bibinfo{author}{Yanase, T.}, \bibinfo{author}{Ohta, T.}, \bibinfo{author}{Koyama, M.}, \bibinfo{year}{2019}.
\newblock \bibinfo{title}{Optuna: A next-generation hyperparameter optimization framework}, in: \bibinfo{booktitle}{Proceedings of the 25th {ACM} {SIGKDD} International Conference on Knowledge Discovery and Data Mining}.
%Type = Article
\bibitem[{Arentze and Timmermans(2004)}]{ALBATROSS}
\bibinfo{author}{Arentze, T.A.}, \bibinfo{author}{Timmermans, H.J.}, \bibinfo{year}{2004}.
\newblock \bibinfo{title}{A learning-based transportation oriented simulation system}.
\newblock \bibinfo{journal}{Transportation Research Part B: Methodological} \bibinfo{volume}{38}, \bibinfo{pages}{613--633}.
\newblock \DOIprefix\doi{https://doi.org/10.1016/j.trb.2002.10.001}.
%Type = Article
\bibitem[{Auld and Mohammadian(2009)}]{ADAPTS}
\bibinfo{author}{Auld, J.}, \bibinfo{author}{Mohammadian, A.}, \bibinfo{year}{2009}.
\newblock \bibinfo{title}{Framework for the development of the agent-based dynamic activity planning and travel scheduling (adapts) model}.
\newblock \bibinfo{journal}{Transportation Letters} \bibinfo{volume}{1}, \bibinfo{pages}{245--255}.
%Type = Misc
\bibitem[{Badu-Marfo et~al.(2020)Badu-Marfo, Farooq and Patterson}]{badumarfo2020differentially}
\bibinfo{author}{Badu-Marfo, G.}, \bibinfo{author}{Farooq, B.}, \bibinfo{author}{Patterson, Z.}, \bibinfo{year}{2020}.
\newblock \bibinfo{title}{A differentially private multi-output deep generative networks approach for activity diary synthesis}.
\newblock \href{http://arxiv.org/abs/2012.14574}{{\tt arXiv:2012.14574}}.
%Type = Article
\bibitem[{Borysov et~al.(2019)Borysov, Rich and Pereira}]{borysovScalablePopulationSynthesis2019}
\bibinfo{author}{Borysov, S.S.}, \bibinfo{author}{Rich, J.}, \bibinfo{author}{Pereira, F.C.}, \bibinfo{year}{2019}.
\newblock \bibinfo{title}{Scalable {{Population Synthesis}} with {{Deep Generative Modeling}}}.
\newblock \bibinfo{journal}{Transportation Research Part C: Emerging Technologies} \bibinfo{volume}{106}, \bibinfo{pages}{73--97}.
\newblock \DOIprefix\doi{10.1016/j.trc.2019.07.006}, \href{http://arxiv.org/abs/1808.06910}{{\tt arXiv:1808.06910}}.
%Type = Article
\bibitem[{Bowman et~al.(2015)Bowman, Vilnis, Vinyals, Dai, J{\'{o}}zefowicz and Bengio}]{sequenceVAE}
\bibinfo{author}{Bowman, S.R.}, \bibinfo{author}{Vilnis, L.}, \bibinfo{author}{Vinyals, O.}, \bibinfo{author}{Dai, A.M.}, \bibinfo{author}{J{\'{o}}zefowicz, R.}, \bibinfo{author}{Bengio, S.}, \bibinfo{year}{2015}.
\newblock \bibinfo{title}{Generating sentences from a continuous space}.
\newblock \bibinfo{journal}{CoRR} \bibinfo{volume}{abs/1511.06349}.
%Type = Article
\bibitem[{Bradley et~al.(2010)Bradley, Bowman and Griesenbeck}]{DaySim}
\bibinfo{author}{Bradley, M.}, \bibinfo{author}{Bowman, J.L.}, \bibinfo{author}{Griesenbeck, B.}, \bibinfo{year}{2010}.
\newblock \bibinfo{title}{Sacsim: An applied activity-based model system with fine-level spatial and temporal resolution}.
\newblock \bibinfo{journal}{Journal of Choice Modelling} \bibinfo{volume}{3}, \bibinfo{pages}{5--31}.
%Type = Article
\bibitem[{Drchal et~al.(2016)Drchal, {\v C}ertick{\'y} and Jakob}]{drchalVALFRAMValidationFramework2016}
\bibinfo{author}{Drchal, J.}, \bibinfo{author}{{\v C}ertick{\'y}, M.}, \bibinfo{author}{Jakob, M.}, \bibinfo{year}{2016}.
\newblock \bibinfo{title}{{{VALFRAM}}: {{Validation Framework}} for {{Activity-Based Models}}}.
\newblock \bibinfo{journal}{Journal of Artificial Societies and Social Simulation} \bibinfo{volume}{19}, \bibinfo{pages}{5}.
%Type = Inproceedings
\bibitem[{Goodfellow et~al.(2014)Goodfellow, Pouget-Abadie, Mirza, Xu, Warde-Farley, Ozair, Courville and Bengio}]{GAN}
\bibinfo{author}{Goodfellow, I.}, \bibinfo{author}{Pouget-Abadie, J.}, \bibinfo{author}{Mirza, M.}, \bibinfo{author}{Xu, B.}, \bibinfo{author}{Warde-Farley, D.}, \bibinfo{author}{Ozair, S.}, \bibinfo{author}{Courville, A.}, \bibinfo{author}{Bengio, Y.}, \bibinfo{year}{2014}.
\newblock \bibinfo{title}{Generative adversarial nets}, in: \bibinfo{editor}{Ghahramani, Z.}, \bibinfo{editor}{Welling, M.}, \bibinfo{editor}{Cortes, C.}, \bibinfo{editor}{Lawrence, N.}, \bibinfo{editor}{Weinberger, K.} (Eds.), \bibinfo{booktitle}{Advances in Neural Information Processing Systems}, \bibinfo{publisher}{Curran Associates, Inc.}
%Type = Inproceedings
\bibitem[{Gregor et~al.(2015)Gregor, Danihelka, Graves, Rezende and Wierstra}]{DRAW}
\bibinfo{author}{Gregor, K.}, \bibinfo{author}{Danihelka, I.}, \bibinfo{author}{Graves, A.}, \bibinfo{author}{Rezende, D.}, \bibinfo{author}{Wierstra, D.}, \bibinfo{year}{2015}.
\newblock \bibinfo{title}{Draw: A recurrent neural network for image generation}, in: \bibinfo{editor}{Bach, F.}, \bibinfo{editor}{Blei, D.} (Eds.), \bibinfo{booktitle}{Proceedings of the 32nd International Conference on Machine Learning}, \bibinfo{publisher}{PMLR}, \bibinfo{address}{Lille, France}. pp. \bibinfo{pages}{1462--1471}.
%Type = Article
\bibitem[{Hafezi et~al.(2021)Hafezi, Daisy, Millward and Liu}]{trees}
\bibinfo{author}{Hafezi, M.H.}, \bibinfo{author}{Daisy, N.S.}, \bibinfo{author}{Millward, H.}, \bibinfo{author}{Liu, L.}, \bibinfo{year}{2021}.
\newblock \bibinfo{title}{Ensemble learning activity scheduler for activity based travel demand models}.
\newblock \bibinfo{journal}{Transportation Research Part C: Emerging Technologies} \bibinfo{volume}{123}, \bibinfo{pages}{102972}.
\newblock \URLprefix \url{https://www.sciencedirect.com/science/article/pii/S0968090X21000097}, \DOIprefix\doi{https://doi.org/10.1016/j.trc.2021.102972}.
%Type = Inproceedings
\bibitem[{Ho et~al.(2020)Ho, Jain and Abbeel}]{diffusion}
\bibinfo{author}{Ho, J.}, \bibinfo{author}{Jain, A.}, \bibinfo{author}{Abbeel, P.}, \bibinfo{year}{2020}.
\newblock \bibinfo{title}{Denoising diffusion probabilistic models}, in: \bibinfo{editor}{Larochelle, H.}, \bibinfo{editor}{Ranzato, M.}, \bibinfo{editor}{Hadsell, R.}, \bibinfo{editor}{Balcan, M.}, \bibinfo{editor}{Lin, H.} (Eds.), \bibinfo{booktitle}{Advances in Neural Information Processing Systems}, \bibinfo{publisher}{Curran Associates, Inc.}. pp. \bibinfo{pages}{6840--6851}.
%Type = Article
\bibitem[{Hochreiter and Schmidhuber(1997)}]{lstm}
\bibinfo{author}{Hochreiter, S.}, \bibinfo{author}{Schmidhuber, J.}, \bibinfo{year}{1997}.
\newblock \bibinfo{title}{{Long Short-Term Memory}}.
\newblock \bibinfo{journal}{Neural Computation} \bibinfo{volume}{9}, \bibinfo{pages}{1735--1780}.
\newblock \DOIprefix\doi{10.1162/neco.1997.9.8.1735}.
%Type = Book
\bibitem[{Horni et~al.(2016)Horni, Nagel and Axhausen}]{MATSim}
\bibinfo{editor}{Horni, A.}, \bibinfo{editor}{Nagel, K.}, \bibinfo{editor}{Axhausen, K.} (Eds.), \bibinfo{year}{2016}.
\newblock \bibinfo{title}{Multi-Agent Transport Simulation MATSim}.
\newblock \bibinfo{publisher}{Ubiquity Press}, \bibinfo{address}{London}.
\newblock \DOIprefix\doi{10.5334/baw}.
%Type = Article
\bibitem[{Khan and Habib(2023)}]{SDS}
\bibinfo{author}{Khan, N.A.}, \bibinfo{author}{Habib, M.A.}, \bibinfo{year}{2023}.
\newblock \bibinfo{title}{Microsimulation of activity generation, activity scheduling and shared travel choices within an activity-based travel demand modelling system}.
\newblock \bibinfo{journal}{Travel Behaviour and Society} \bibinfo{volume}{32}, \bibinfo{pages}{100590}.
\newblock \DOIprefix\doi{https://doi.org/10.1016/j.tbs.2023.100590}.
%Type = Article
\bibitem[{Kim and Bansal(2023)}]{kimDeepGenerativeModel2023a}
\bibinfo{author}{Kim, E.J.}, \bibinfo{author}{Bansal, P.}, \bibinfo{year}{2023}.
\newblock \bibinfo{title}{A deep generative model for feasible and diverse population synthesis}.
\newblock \bibinfo{journal}{Transportation Research Part C: Emerging Technologies} \bibinfo{volume}{148}, \bibinfo{pages}{104053}.
\newblock \DOIprefix\doi{10.1016/j.trc.2023.104053}.
%Type = Article
\bibitem[{Kim et~al.(2022)Kim, Kim and Sohn}]{tripCGAN}
\bibinfo{author}{Kim, E.J.}, \bibinfo{author}{Kim, D.K.}, \bibinfo{author}{Sohn, K.}, \bibinfo{year}{2022}.
\newblock \bibinfo{title}{Imputing qualitative attributes for trip chains extracted from smart card data using a conditional generative adversarial network}.
\newblock \bibinfo{journal}{Transportation Research Part C: Emerging Technologies} \bibinfo{volume}{137}, \bibinfo{pages}{103616}.
\newblock \DOIprefix\doi{10.1016/j.trc.2022.103616}.
%Type = Misc
\bibitem[{Kingma and Welling(2013)}]{vae}
\bibinfo{author}{Kingma, D.P.}, \bibinfo{author}{Welling, M.}, \bibinfo{year}{2013}.
\newblock \bibinfo{title}{Auto-{{Encoding Variational Bayes}}}.
\newblock \DOIprefix\doi{10.48550/arXiv.1312.6114}, \href{http://arxiv.org/abs/1312.6114}{{\tt arXiv:1312.6114}}.
%Type = Article
\bibitem[{Koushik et~al.(2023)Koushik, Manoj, Nezamuddin and Prathosh}]{koushikActivityScheduleModeling2023}
\bibinfo{author}{Koushik, A.}, \bibinfo{author}{Manoj, M.}, \bibinfo{author}{Nezamuddin, N.}, \bibinfo{author}{Prathosh, {\relax AP}.}, \bibinfo{year}{2023}.
\newblock \bibinfo{title}{Activity {{Schedule Modeling Using Machine Learning}}}.
\newblock \bibinfo{journal}{Transportation Research Record} \bibinfo{volume}{2677}, \bibinfo{pages}{1--23}.
\newblock \DOIprefix\doi{10.1177/03611981231155426}.
%Type = Misc
\bibitem[{Liao et~al.(2024)Liao, Jiang, He, Liu, Kuai and Ma}]{deepactivitymodelgenerative}
\bibinfo{author}{Liao, X.}, \bibinfo{author}{Jiang, Q.}, \bibinfo{author}{He, B.Y.}, \bibinfo{author}{Liu, Y.}, \bibinfo{author}{Kuai, C.}, \bibinfo{author}{Ma, J.}, \bibinfo{year}{2024}.
\newblock \bibinfo{title}{Deep activity model: A generative approach for human mobility pattern synthesis}.
\newblock \URLprefix \url{https://arxiv.org/abs/2405.17468}, \href{http://arxiv.org/abs/2405.17468}{{\tt arXiv:2405.17468}}.
%Type = Article
\bibitem[{Liu et~al.(2015)Liu, Janssens, Cui, Wets and Cools}]{pHMM}
\bibinfo{author}{Liu, F.}, \bibinfo{author}{Janssens, D.}, \bibinfo{author}{Cui, J.}, \bibinfo{author}{Wets, G.}, \bibinfo{author}{Cools, M.}, \bibinfo{year}{2015}.
\newblock \bibinfo{title}{Characterizing activity sequences using profile hidden markov models}.
\newblock \bibinfo{journal}{Expert Systems with Applications} \bibinfo{volume}{42}, \bibinfo{pages}{5705--5722}.
\newblock \DOIprefix\doi{https://doi.org/10.1016/j.eswa.2015.02.057}.
%Type = Inproceedings
\bibitem[{Lucas et~al.(2019)Lucas, Shmelkov, Alahari, Schmid and Verbeek}]{GANdensity}
\bibinfo{author}{Lucas, T.}, \bibinfo{author}{Shmelkov, K.}, \bibinfo{author}{Alahari, K.}, \bibinfo{author}{Schmid, C.}, \bibinfo{author}{Verbeek, J.}, \bibinfo{year}{2019}.
\newblock \bibinfo{title}{Adaptive density estimation for generative models}, in: \bibinfo{editor}{Wallach, H.}, \bibinfo{editor}{Larochelle, H.}, \bibinfo{editor}{Beygelzimer, A.}, \bibinfo{editor}{d\textquotesingle Alch\'{e}-Buc, F.}, \bibinfo{editor}{Fox, E.}, \bibinfo{editor}{Garnett, R.} (Eds.), \bibinfo{booktitle}{Advances in Neural Information Processing Systems}, \bibinfo{publisher}{Curran Associates, Inc.}
\newblock \URLprefix \url{https://proceedings.neurips.cc/paper_files/paper/2019/file/959ab9a0695c467e7caf75431a872e5c-Paper.pdf}.
%Type = Misc
\bibitem[{Manser et~al.(2021)Manser, Haering, Hillel, Pougala, Krueger and Bierlaire}]{Manser2021ResolvingTS}
\bibinfo{author}{Manser, P.}, \bibinfo{author}{Haering, T.}, \bibinfo{author}{Hillel, T.}, \bibinfo{author}{Pougala, J.}, \bibinfo{author}{Krueger, R.}, \bibinfo{author}{Bierlaire, M.}, \bibinfo{year}{2021}.
\newblock \bibinfo{title}{Resolving temporal scheduling conflicts in activity-based modelling}.
\newblock \URLprefix \url{https://transp-or.epfl.ch/documents/technicalReports/ManserEtAl2021.pdf}.
%Type = Article
\bibitem[{Miller and Roorda(2003)}]{TASHA}
\bibinfo{author}{Miller, E.J.}, \bibinfo{author}{Roorda, M.J.}, \bibinfo{year}{2003}.
\newblock \bibinfo{title}{Prototype model of household activity-travel scheduling}.
\newblock \bibinfo{journal}{Transportation Research Record} \bibinfo{volume}{1831}, \bibinfo{pages}{114--121}.
%Type = Article
\bibitem[{Miller et~al.(2005)Miller, Roorda and Carrasco}]{miller2005tour}
\bibinfo{author}{Miller, E.J.}, \bibinfo{author}{Roorda, M.J.}, \bibinfo{author}{Carrasco, J.A.}, \bibinfo{year}{2005}.
\newblock \bibinfo{title}{A tour-based model of travel mode choice}.
\newblock \bibinfo{journal}{Transportation} \bibinfo{volume}{32}, \bibinfo{pages}{399--422}.
%Type = Inproceedings
\bibitem[{Naeem et~al.(2020)Naeem, Oh, Uh, Choi and Yoo}]{fidelity}
\bibinfo{author}{Naeem, M.F.}, \bibinfo{author}{Oh, S.J.}, \bibinfo{author}{Uh, Y.}, \bibinfo{author}{Choi, Y.}, \bibinfo{author}{Yoo, J.}, \bibinfo{year}{2020}.
\newblock \bibinfo{title}{Reliable fidelity and diversity metrics for generative models}, in: \bibinfo{editor}{III, H.D.}, \bibinfo{editor}{Singh, A.} (Eds.), \bibinfo{booktitle}{Proceedings of the 37th International Conference on Machine Learning}, \bibinfo{publisher}{PMLR}. pp. \bibinfo{pages}{7176--7185}.
\newblock \URLprefix \url{https://proceedings.mlr.press/v119/naeem20a.html}.
%Type = Article
\bibitem[{Nayak and Pandit(2023)}]{dap_ml}
\bibinfo{author}{Nayak, S.}, \bibinfo{author}{Pandit, D.}, \bibinfo{year}{2023}.
\newblock \bibinfo{title}{A joint and simultaneous prediction framework of weekday and weekend daily-activity travel pattern using conditional dependency networks}.
\newblock \bibinfo{journal}{Travel Behaviour and Society} \bibinfo{volume}{32}, \bibinfo{pages}{100595}.
\newblock \URLprefix \url{https://www.sciencedirect.com/science/article/pii/S2214367X23000467}, \DOIprefix\doi{https://doi.org/10.1016/j.tbs.2023.100595}.
%Type = Inproceedings
\bibitem[{van~den Oord et~al.(2017)van~den Oord, Vinyals and kavukcuoglu}]{VQVAE}
\bibinfo{author}{van~den Oord, A.}, \bibinfo{author}{Vinyals, O.}, \bibinfo{author}{kavukcuoglu, k.}, \bibinfo{year}{2017}.
\newblock \bibinfo{title}{Neural discrete representation learning}, in: \bibinfo{editor}{Guyon, I.}, \bibinfo{editor}{Luxburg, U.V.}, \bibinfo{editor}{Bengio, S.}, \bibinfo{editor}{Wallach, H.}, \bibinfo{editor}{Fergus, R.}, \bibinfo{editor}{Vishwanathan, S.}, \bibinfo{editor}{Garnett, R.} (Eds.), \bibinfo{booktitle}{Advances in Neural Information Processing Systems}, \bibinfo{publisher}{Curran Associates, Inc.}
%Type = Article
\bibitem[{Oord et~al.(2016)Oord, Kalchbrenner, Espeholt, Kavukcuoglu, Vinyals and Graves}]{PixelCNN}
\bibinfo{author}{Oord, A.V.D.}, \bibinfo{author}{Kalchbrenner, N.}, \bibinfo{author}{Espeholt, L.}, \bibinfo{author}{Kavukcuoglu, K.}, \bibinfo{author}{Vinyals, O.}, \bibinfo{author}{Graves, A.}, \bibinfo{year}{2016}.
\newblock \bibinfo{title}{Conditional image generation with pixelcnn decoders}.
\newblock \bibinfo{journal}{ArXiv} \bibinfo{volume}{abs/1606.05328}.
%Type = Article
\bibitem[{Pendyala et~al.(2005)Pendyala, Kitamura, Kikuchi, Yamamoto and Fujii}]{FAMOS}
\bibinfo{author}{Pendyala, R.M.}, \bibinfo{author}{Kitamura, R.}, \bibinfo{author}{Kikuchi, A.}, \bibinfo{author}{Yamamoto, T.}, \bibinfo{author}{Fujii, S.}, \bibinfo{year}{2005}.
\newblock \bibinfo{title}{Florida activity mobility simulator: overview and preliminary validation results}.
\newblock \bibinfo{journal}{Transportation Research Record} \bibinfo{volume}{1921}, \bibinfo{pages}{123--130}.
%Type = Article
\bibitem[{Pougala et~al.(2023)Pougala, Hillel and Bierlaire}]{POUGALA2023104291}
\bibinfo{author}{Pougala, J.}, \bibinfo{author}{Hillel, T.}, \bibinfo{author}{Bierlaire, M.}, \bibinfo{year}{2023}.
\newblock \bibinfo{title}{Oasis: Optimisation-based activity scheduling with integrated simultaneous choice dimensions}.
\newblock \bibinfo{journal}{Transportation Research Part C: Emerging Technologies} \bibinfo{volume}{155}, \bibinfo{pages}{104291}.
\newblock \DOIprefix\doi{https://doi.org/10.1016/j.trc.2023.104291}.
%Type = Inproceedings
\bibitem[{Rezende and Mohamed(2015)}]{norming_flows}
\bibinfo{author}{Rezende, D.}, \bibinfo{author}{Mohamed, S.}, \bibinfo{year}{2015}.
\newblock \bibinfo{title}{Variational inference with normalizing flows}, in: \bibinfo{editor}{Bach, F.}, \bibinfo{editor}{Blei, D.} (Eds.), \bibinfo{booktitle}{Proceedings of the 32nd International Conference on Machine Learning}, \bibinfo{publisher}{PMLR}, \bibinfo{address}{Lille, France}. pp. \bibinfo{pages}{1530--1538}.
%Type = Inproceedings
\bibitem[{Rezende et~al.(2014)Rezende, Mohamed and Wierstra}]{rezende2014stochastic}
\bibinfo{author}{Rezende, D.J.}, \bibinfo{author}{Mohamed, S.}, \bibinfo{author}{Wierstra, D.}, \bibinfo{year}{2014}.
\newblock \bibinfo{title}{Stochastic backpropagation and approximate inference in deep generative models}, in: \bibinfo{booktitle}{International conference on machine learning}, \bibinfo{organization}{PMLR}. pp. \bibinfo{pages}{1278--1286}.
%Type = Inproceedings
\bibitem[{Roberts et~al.(2018)Roberts, Engel, Raffel, Hawthorne and Eck}]{musicVAE}
\bibinfo{author}{Roberts, A.}, \bibinfo{author}{Engel, J.}, \bibinfo{author}{Raffel, C.}, \bibinfo{author}{Hawthorne, C.}, \bibinfo{author}{Eck, D.}, \bibinfo{year}{2018}.
\newblock \bibinfo{title}{A hierarchical latent vector model for learning long-term structure in music}, in: \bibinfo{editor}{Dy, J.}, \bibinfo{editor}{Krause, A.} (Eds.), \bibinfo{booktitle}{Proceedings of the 35th International Conference on Machine Learning}, \bibinfo{publisher}{PMLR}. pp. \bibinfo{pages}{4364--4373}.
%Type = Inproceedings
\bibitem[{Sajjadi et~al.(2018)Sajjadi, Bachem, Lucic, Bousquet and Gelly}]{precandrecal}
\bibinfo{author}{Sajjadi, M.S.M.}, \bibinfo{author}{Bachem, O.}, \bibinfo{author}{Lucic, M.}, \bibinfo{author}{Bousquet, O.}, \bibinfo{author}{Gelly, S.}, \bibinfo{year}{2018}.
\newblock \bibinfo{title}{Assessing generative models via precision and recall}, in: \bibinfo{editor}{Bengio, S.}, \bibinfo{editor}{Wallach, H.}, \bibinfo{editor}{Larochelle, H.}, \bibinfo{editor}{Grauman, K.}, \bibinfo{editor}{Cesa-Bianchi, N.}, \bibinfo{editor}{Garnett, R.} (Eds.), \bibinfo{booktitle}{Advances in Neural Information Processing Systems}, \bibinfo{publisher}{Curran Associates, Inc.}
\newblock \URLprefix \url{https://proceedings.neurips.cc/paper_files/paper/2018/file/f7696a9b362ac5a51c3dc8f098b73923-Paper.pdf}.
%Type = Techreport
\bibitem[{Sener et~al.(2006)Sener, Bhat, Copperman, Srinivasan, Guo, Pinjari and Eluru}]{CEMDAP}
\bibinfo{author}{Sener, I.N.}, \bibinfo{author}{Bhat, C.R.}, \bibinfo{author}{Copperman, R.}, \bibinfo{author}{Srinivasan, S.}, \bibinfo{author}{Guo, J.Y.}, \bibinfo{author}{Pinjari, A.}, \bibinfo{author}{Eluru, N.}, \bibinfo{year}{2006}.
\newblock \bibinfo{title}{Activity-based travel-demand analysis for metropolitan areas in Texas: CEMDAP models, framework, software architecture and application results}.
\newblock \bibinfo{type}{Technical Report}. Midwest Regional University Transportation Center.
%Type = Article
\bibitem[{Shone et~al.(2024)Shone, Chatziioannou, Pickering, Kozlowska and Fitzmaurice}]{PAM}
\bibinfo{author}{Shone, F.}, \bibinfo{author}{Chatziioannou, T.}, \bibinfo{author}{Pickering, B.}, \bibinfo{author}{Kozlowska, K.}, \bibinfo{author}{Fitzmaurice, M.}, \bibinfo{year}{2024}.
\newblock \bibinfo{title}{{PAM: Population Activity Modeller}}.
\newblock \bibinfo{journal}{Journal of Open Source Software} \bibinfo{volume}{9}, \bibinfo{pages}{6097}.
\newblock \DOIprefix\doi{10.21105/joss.06097}.
%Type = Inproceedings
\bibitem[{Theis et~al.(2016)Theis, van~den Oord and Bethge}]{deep_eval}
\bibinfo{author}{Theis, L.}, \bibinfo{author}{van~den Oord, A.}, \bibinfo{author}{Bethge, M.}, \bibinfo{year}{2016}.
\newblock \bibinfo{title}{A note on the evaluation of generative models}, in: \bibinfo{editor}{Bengio, Y.}, \bibinfo{editor}{LeCun, Y.} (Eds.), \bibinfo{booktitle}{4th International Conference on Learning Representations, {ICLR} 2016, San Juan, Puerto Rico, May 2-4, 2016, Conference Track Proceedings}.
\newblock \URLprefix \url{http://arxiv.org/abs/1511.01844}.
%Type = Inproceedings
\bibitem[{Vaswani et~al.(2017)Vaswani, Shazeer, Parmar, Uszkoreit, Jones, Gomez, Kaiser and Polosukhin}]{AIAYN}
\bibinfo{author}{Vaswani, A.}, \bibinfo{author}{Shazeer, N.}, \bibinfo{author}{Parmar, N.}, \bibinfo{author}{Uszkoreit, J.}, \bibinfo{author}{Jones, L.}, \bibinfo{author}{Gomez, A.N.}, \bibinfo{author}{Kaiser, L.u.}, \bibinfo{author}{Polosukhin, I.}, \bibinfo{year}{2017}.
\newblock \bibinfo{title}{Attention is all you need}, in: \bibinfo{editor}{Guyon, I.}, \bibinfo{editor}{Luxburg, U.V.}, \bibinfo{editor}{Bengio, S.}, \bibinfo{editor}{Wallach, H.}, \bibinfo{editor}{Fergus, R.}, \bibinfo{editor}{Vishwanathan, S.}, \bibinfo{editor}{Garnett, R.} (Eds.), \bibinfo{booktitle}{Advances in Neural Information Processing Systems}, \bibinfo{publisher}{Curran Associates, Inc.}

\end{thebibliography}





% \end{thebibliography}
\end{document}